\gdef\isdm{1}
\ifdefined\isdm
\documentclass[11pt, a4paper]{deepmind}
\renewcommand{\today}{7 October 2020}
\else
\documentclass{article} 
\fi

\ifdefined\isdm
\usepackage[authoryear, sort&compress, round]{natbib}
\else
\usepackage[table]{xcolor}
\usepackage{iclr2021_conference,times}
\fi
\usepackage{verbatim}
\usepackage{hyperref}
\usepackage{url}
\usepackage{graphicx}
\usepackage{hyperref}
\usepackage{url}
\usepackage{wrapfig}
\usepackage{glossaries}
\usepackage{xspace}
\usepackage{subcaption}
\usepackage{multirow}
\usepackage{tocloft}
\usepackage{etoc}
\usepackage{floatflt}

\usepackage{amsmath,amsfonts,bm}

\def\eqref#1{equation~\ref{#1}}

\def\1{\bm{1}}

\def\vone{{\bm{1}}}

\def\vtheta{{\bm{\theta}}}
\def\vdelta{{\bm{\delta}}}
\def\va{{\bm{a}}}

\def\vx{{\bm{x}}}

\DeclareMathAlphabet{\mathsfit}{\encodingdefault}{\sfdefault}{m}{sl}
\SetMathAlphabet{\mathsfit}{bold}{\encodingdefault}{\sfdefault}{bx}{n}

\def\sA{{\mathbb{A}}}

\def\sS{{\mathbb{S}}}

\newcommand{\E}{\mathbb{E}}

\DeclareMathOperator*{\argmax}{arg\,max}
\DeclareMathOperator*{\argmin}{arg\,min}

\DeclareMathOperator{\sign}{sign}

\DeclareMathOperator*{\maximize}{max}

\DeclareMathOperator{\proj}{proj}

\newcommand{\cifar}{\textsc{Cifar-10}\xspace}
\newcommand{\mnist}{\textsc{Mnist}\xspace}
\newcommand{\svhn}{\textsc{Svhn}\xspace}
\newcommand{\cifarh}{\textsc{Cifar-100}\xspace}
\newcommand{\tinyimages}{\textsc{80M-Ti}\xspace}
\newcommand{\imagenet}{\textsc{ImageNet}\xspace}
\newcommand{\linf}{\ensuremath{\ell_\infty}\xspace}
\newcommand{\ltwo}{\ensuremath{\ell_2}\xspace}
\newcommand{\autoattack}{\textsc{AutoAttack}\xspace}
\newcommand{\autopgd}{\textsc{AutoPgd}\xspace}
\newcommand{\multitargeted}{\textsc{MultiTargeted}\xspace}
\newcommand{\kl}{\ensuremath{D_{\textrm{KL}}}\xspace}

\newcommand{\pgd}[1]{\textsc{Pgd}\textsuperscript{$#1$}\xspace}
\newcommand{\xent}{l_{\textrm{xent}}}
\newacronym{fgsm}{FGSM}{Fast Gradient Sign Method}
\newacronym{pgd}{PGD}{Projected Gradient Descent}
\newacronym{bim}{BIM}{Basic Iterative Method}
\newacronym{wrn}{\textsc{Wrn}}{Wide ResNet}
\newacronym{sgd}{SGD}{Stochastic Gradient Descent}
\newcommand{\wrn}{\gls*{wrn}\xspace}
\newcommand{\swish}{\texttt{Swish}/\texttt{SiLU}\xspace}

\renewcommand\UrlFont{\fontsize{8}{10}\selectfont}

\definecolor{header}{gray}{0.9}
\definecolor{subheader}{rgb}{0.63, 0.79, 0.95}

\newcommand{\Tstrut}{\rule{0pt}{2.6ex}}
\newcommand{\Bstrut}{\rule[-0.9ex]{0pt}{0pt}}
\newcommand{\TBstrut}{\Tstrut\Bstrut}

\newcommand{\squishlist}{
   \begin{list}{$\bullet$}
    { \setlength{\itemsep}{0pt}      \setlength{\parsep}{3pt}
      \setlength{\topsep}{3pt}       \setlength{\partopsep}{0pt}
      \setlength{\leftmargin}{1.5em} \setlength{\labelwidth}{1em}
      \setlength{\labelsep}{0.5em} } }
\newcommand{\squishend}{
    \end{list}  }
    
\graphicspath{{images/}}

\title{Uncovering the Limits of Adversarial Training against Norm-Bounded Adversarial Examples}

\ifdefined\isdm
\author[1]{Sven Gowal}
\author[1]{Chongli Qin}
\author[1]{Jonathan Uesato}
\author[1]{Timothy Mann}
\author[1]{Pushmeet Kohli}
\affil[1]{DeepMind}
\correspondingauthor{sgowal@deepmind.com}
\else
\author{Sven Gowal, Chongli Qin, Jonathan Uesato, Timothy Mann \& Pushmeet Kohli \\
DeepMind \\
London, United Kingdom \\
\texttt{sgowal@google.com}}
\fi

\def\myabstract{
Adversarial training and its variants have become de facto standards for learning robust deep neural networks.
In this paper, we explore the landscape around adversarial training in a bid to uncover its limits.
We systematically study the effect of different training losses, model sizes, activation functions, the addition of unlabeled data (through pseudo-labeling) and other factors on adversarial robustness.
We discover that it is possible to train robust models that go well beyond state-of-the-art results by combining larger models, Swish/SiLU activations and model weight averaging.
We demonstrate large improvements on \cifar and \cifarh against \linf and \ltwo norm-bounded perturbations of size $8/255$ and $128/255$, respectively.
In the setting with additional unlabeled data, we obtain an accuracy under attack of 65.88\% against \linf perturbations of size $8/255$ on \cifar (+6.35\% with respect to prior art). 
Without additional data, we obtain an accuracy under attack of 57.20\% (+3.46\%).  
To test the generality of our findings and without any additional modifications, we obtain an accuracy under attack of 80.53\% (+7.62\%) against \ltwo perturbations of size $128/255$ on \cifar, and of 36.88\% (+8.46\%) against \linf perturbations of size $8/255$ on \cifarh.  
All models are available at \url{https://github.com/deepmind/deepmind-research/tree/master/adversarial_robustness}.
}

\ifdefined\isdm
\begin{abstract}
\myabstract
\end{abstract}
\else
\iclrfinalcopy 
\fi
\begin{document}

\maketitle

\ifdefined\isdm\else
\begin{abstract}
\myabstract
\end{abstract}
\fi


\section{Introduction}

\begin{wrapfigure}{r}{0.5\textwidth}
\begin{center}
\vspace{-1cm}
\includegraphics[width=.45\textwidth]{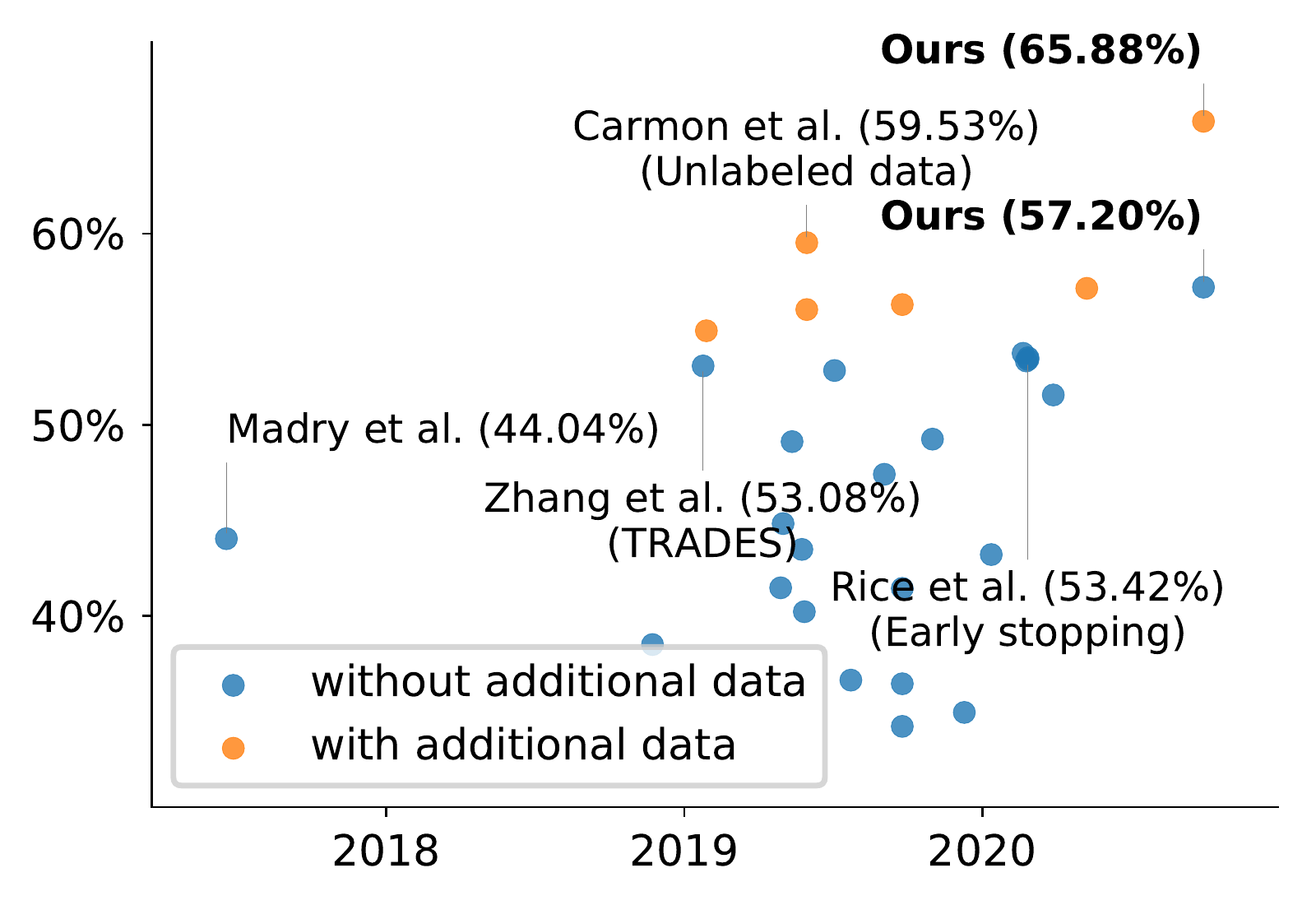}
\end{center}
\caption{Accuracy of various models ordered by publication date against \autoattack~\citep{croce_reliable_2020} on \cifar with \linf perturbations of size $8/255$. Our newest models (on the far right) improve robust accuracy by +3.46\% without additional data and by +6.35\% when using additional unlabeled data. \label{fig:history}}
\end{wrapfigure}

Neural networks are being deployed in a wide variety of applications with great success \citep{goodfellow_deep_2016,krizhevsky_imagenet_2012,hinton_deep_2012}.
As neural networks tackle challenges ranging from ranking content on the web~\citep{covington_deep_2016} to autonomous driving~\citep{bojarski_end_2016} via medical diagnostics~\citep{fauw_clinically_2018}, it has become increasingly important to ensure that deployed models are robust and generalize to various input perturbations.
Despite their success, neural networks are not intrinsically robust.
In particular, the addition of small but carefully chosen deviations to the input, called adversarial perturbations, can cause the neural network to make incorrect predictions with high confidence \citep{carlini_adversarial_2017,goodfellow_explaining_2014,kurakin_adversarial_2016,szegedy_intriguing_2013}.
Starting with \citet{szegedy_intriguing_2013}, there has been a lot of work on understanding and generating adversarial perturbations \citep{carlini_towards_2017,athalye_synthesizing_2017}, and on building models that are robust to such perturbations \citep{goodfellow_explaining_2014,papernot_distillation_2015,madry_towards_2017,kannan_adversarial_2018}.
Robust optimization techniques, like the one developed by \citet{madry_towards_2017}, learn robust models by finding worst-case adversarial examples (by running an inner optimization procedure) at each training step and adding them to the training data.
This technique has proven to be effective and is now widely adopted. 

Since \citet{madry_towards_2017}, various modifications to their algorithm have been proposed \citep{xie_feature_2018,pang_boosting_2020,huang_self-adaptive_2020,qin_adversarial_2019,zoran_towards_2019,andriushchenko_understanding_2020}.
We highlight the work by \citet{zhang_theoretically_2019} who proposed TRADES which balances the trade-off between standard and robust accuracy, and the work by \citet{carmon_unlabeled_2019,uesato_are_2019,najafi_robustness_2019,zhai_adversarially_2019} who simultaneously proposed the use of additional unlabeled data in this context.
As shown in \autoref{fig:history}, despite this flurry of activity, progress over the past two years has been slow.
In a similar spirit to works exploring transfer learning \citep{raffel2019exploring} or recurrent architectures \citep{jozefowicz}, we perform a systematic study around the landscape of adversarial training in a bid to discover its limits. 
Concretely, we study the effects of \textit{(i)} the objective used in the inner and outer optimization procedures, \textit{(ii)} the quantity and quality of additional unlabeled data, \textit{(iii)} the model size, as well as \textit{(iv)} other factors (such as the use of model weight averaging).
In total we trained more than 150 adversarially robust models and dissected each of them to uncover new ideas that could improve adversarial training and our understanding of robustness.
Here is a non-exhaustive list highlighting our findings (in no specific order): 
\squishlist
\item TRADES~\citep{zhang_theoretically_2019} combined with early stopping outperforms regular adversarial training (as proposed by \citealp{madry_towards_2017}). This is in contrast to the observations made by \citet{rice_overfitting_2020} (\autoref{sec:outer_minimization}).
\item As observed by \citet{madry_towards_2017,xie_intriguing_2019,uesato_are_2019}, increasing the capacity of models improves robustness (\autoref{sec:modelsize}).
\item The choice of activation function matters. Similar to observations made by \citet{xie_smooth_2020}, we found that \swish~\citep{hendrycks2016gaussian,ramachandran2017searching,elfwing2017sigmoidweighted} performs best. However, in contrast to~\citeauthor{xie_smooth_2020}, we found that other ``smooth'' activation functions do not necessarily improve robustness (\autoref{sec:activations}).
\item The way in which additional unlabeled data is extracted from the 80 Million Tiny Images dataset (\tinyimages)~\citep{80m} and used during training can have a significant impact (\autoref{sec:unlabeled}).
\item Model weight averaging (WA)~\citep{izmailov_averaging_2018} consistently provides a boost in robustness. In the setting without additional unlabeled data, WA provides improvements as large as those provided by TRADES on top of classical adversarial training (\autoref{sec:ema}).
\squishend

These findings result in a suite of models that significantly improve on the state-of-the-art against \linf and \ltwo norm-bounded perturbations on \cifar, and against \linf norm-bounded perturbations on \cifarh and \mnist (as measured by \autoattack):\footnote{\url{https://github.com/fra31/auto-attack}}
\squishlist
\item On \cifar against \linf norm-bounded perturbations of size $\epsilon = 8/255$, we train models with 65.88\% and 57.20\% robust accuracy with and without additional unlabeled data, respectively (at the time of writing, prior art was 59.53\% and 53.74\% in both settings).
\item With the same settings used for \cifar against \linf norm-bounded perturbations, on \cifar against \ltwo norm-bounded perturbations of size $\epsilon = 128/255$, we train models with 80.53\% and 74.50\% robust accuracy with and without additional unlabeled data, respectively (prior art was 72.91\% and 69.24\%).
\item On \cifarh against \linf norm-bounded perturbations of size $\epsilon = 8/255$, we train models with 36.88\% and 30.03\% robust accuracy with and without additional unlabeled data, respectively  (prior art was 28.42\% and 18.95\%).
\squishend
In this study, we aim to find the current limits of adversarial robustness.
What we found was that an accumulation of small factors can significantly improve upon the state-of-the-art when combined.
As fundamentally new techniques might be needed, it is important to understand the limitations of current approaches.
We hope that this new set of baselines can further new understanding about adversarial robustness.

\section{Background}

\subsection{Context}
\label{sec:related_work}

\paragraph{Adversarial attacks.}

Since \citet{szegedy_intriguing_2013} observed that neural networks which achieve high accuracy on test data are highly vulnerable to adversarial examples, the art of crafting increasingly sophisticated adversarial examples has received a lot of attention.
\citet{goodfellow_explaining_2014} proposed the \gls*{fgsm} which generates adversarial examples with a single normalized gradient step.
It was followed by R+\gls*{fgsm} \citep{tramer_ensemble_2017}, which adds a randomization step, and the \gls*{bim} \citep{kurakin_adversarial_2016}, which takes multiple smaller gradient steps.
These are often grouped under the term \gls*{pgd} which usually refers to the optimization procedure used to search norm-bounded perturbations.

\vspace{-.2cm}
\paragraph{Adversarial training as a defense.}

The adversarial training procedure which feeds adversarially perturbed examples back into the training data is widely regarded as one of the most successful method to train robust deep neural networks.
Its classical version detailed by \citet{madry_towards_2017} has been augmented in different ways -- with changes in the attack procedure (e.g., by incorporating momentum; \citealp{dong_boosting_2017}), loss function (e.g., logit pairing; \citealp{mosbach_logit_2018}) or model architecture (e.g., feature denoising; \citealp{xie_feature_2018}).
Other notable works include the work by \citet{zhang_theoretically_2019} who proposed TRADES which balances the trade-off between standard and robust accuracy, and the work by \citet{wang_improving_2020} who proposed MART which also addresses this trade-off by using boosted loss functions.
Both works achieved state-of-the-art performance against \linf norm-bounded perturbations on \cifar.
The work from \cite{rice_overfitting_2020} stood out as a study on robust overfitting which demonstrated that improvements similar to TRADES and MART could be obtained more easily using classical adversarial training with early stopping.
This study revealed that there is much we do not yet understand about adversarial training, and serves as one of our motivations for performing a holistic analysis of different aspects of adversarial training.
So far, to the best our knowledge, there has been no systematic study of adversarial training.

\vspace{-.2cm}
\paragraph{Other defenses.}

Many other alternative defenses are not covered in the scope of this paper.
They range from preprocessing techniques \citep{guo_countering_2017, buckman_thermometer_2018} to detection algorithms \citep{metzen_detecting_2017, feinman_detecting_2017}, and also include the definition of new regularizers \citep{moosavi-dezfooli_robustness_2018, xiao_training_2018, qin_adversarial_2019}.
The difficulty of adversarial evaluation also drove the need for \emph{certified defenses}~\citep{wong_scaling_2018, mirman_differentiable_2018, gowal_effectiveness_2018, zhang_towards_2019, cohen_certified_2019, salman_provably_2019}, but the guarantees that these techniques provide do not yet match the empirical robustness obtained through adversarial training.

\vspace{-.2cm}
\paragraph{Adversarial evaluation.}
It is worth noting that many of the defense strategies proposed in the literature \citep{papernot_distillation_2015, lu_no_2017, kannan_adversarial_2018, tao_attacks_2018, zhang_defense_2019} were broken by stronger adversaries \citep{carlini_defensive_2016, carlini_towards_2017, athalye_synthesizing_2017, engstrom_evaluating_2018, carlini_is_2019, uesato_adversarial_2018, athalye_obfuscated_2018}.
Hence, the robust accuracy obtained under different evaluation protocols cannot be easily compared and care has to be taken to make sure the evaluation is as strong as is possible \citep{carlini_evaluating_2019}.
In this manuscript, we evaluate each model against two of the strongest adversarial attacks, \autoattack and \multitargeted, developed by \citet{croce_reliable_2020} and \citet{gowal_alternative_2019}, respectively.

\subsection{Adversarial training}
\label{sec:at}

\citet{madry_towards_2017} formulate a saddle point problem whose goal is to find model parameters $\vtheta$ that minimize the adversarial risk:
\begin{equation}
\underbrace{\argmin_\vtheta \E_{(\vx,y) \sim \mathcal{D}} \underbrace{\left[ \maximize_{\vdelta \in \sS} l(f(\vx + \vdelta; \vtheta), y) \right]}_{\mbox{inner maximization}}}_{\mbox{outer minimization}}
\label{eq:adversarial_risk}
\end{equation}
where $\mathcal{D}$ is a data distribution over pairs of examples $\vx$ and corresponding labels $y$, $f(\cdot; \vtheta)$ is a model parametrized by $\vtheta$, $l$ is a suitable loss function (such as the $0-1$ loss in the context of classification tasks), and $\sS$ defines the set of allowed perturbations (i.e., the adversarial input set or threat model).
For $\ell_p$ norm-bounded perturbations of size $\epsilon$, the adversarial set is defined as $\sS_p = \{ \vdelta ~|~ \| \vdelta \|_p < \epsilon \}$.
Hence, for \linf norm-bounded perturbations $\sS = \sS_\infty$ and for \ltwo norm-bounded perturbations $\sS = \sS_2$

\paragraph{Inner maximization.}
As finding the optimum of the inner maximization problem is NP-hard, several methods (also known as ``attacks'') have been proposed to approximate its solution.
\citet{madry_towards_2017} use \gls*{pgd},\footnote{There exists a few variants of PGD which normalize the gradient step differently (e.g., using an \ltwo normalization for \ltwo norm-bounded perturbations).} which replaces the non-differentiable $0-1$ loss $l$ with the cross-entropy loss $\xent$ and computes an adversarial perturbation $\hat{\vdelta} = \vdelta^{(K)}$ in $K$ gradient ascent steps of size $\alpha$ as
\begin{equation}
\vdelta^{(k+1)} \gets \proj_{\sS} \left( \vdelta^{(k)} + \alpha \sign \left(\nabla_{\vdelta^{(t)}} \xent(f(\vx + \vdelta^{(k)}; \vtheta), y) \right)\right) \label{eq:bim}
\end{equation}
where $\vdelta^{(0)}$ is chosen at random within $\sS$, and where $\proj_{\sA}(\va)$ projects a point $\va$ back onto a set $\sA$, $\proj_{\sA}(\va) = \mathrm{argmin}_{\va' \in \sA} \|\va - \va'\|_2$.
We will refer to this inner optimization procedure with $K$ steps as \pgd{K}.

\paragraph{Outer minimization.}

For each example $\vx$ with label $y$, adversarial training minimizes the loss given by
\begin{equation}
\mathcal{L}_\vtheta^\textrm{AT} = \xent(f(\vx + \hat{\vdelta}; \vtheta), y) \approx \maximize_{\vdelta \in \sS} \xent(f(\vx + \vdelta; \vtheta), y) \label{eq:at_loss}
\end{equation}
where $\hat{\vdelta}$ is given by \autoref{eq:bim} and $\xent$ is the softmax cross-entropy loss.

\section{Setup and implementation details}

Initially, we focus on robustness to \linf norm-bounded perturbations on \cifar of size $\epsilon = 8/255$.
\autoref{sec:combined} combines individual components to surpass the state-of-the-art and tests their generality against other datasets (i.e., \cifarh and \mnist) and another threat model (i.e., \ltwo norm-bounded perturbations of size $\epsilon = 128/255$).
First, we describe the experimental setup.

\subsection{Setup}
\label{sec:setup}

\paragraph{Architecture.}

For consistency with prior work on adversarial robustness~\citep{madry_towards_2017,rice_overfitting_2020,zhang_theoretically_2019,uesato_are_2019}, we use \glspl{wrn}~\citep{he2015deep,zagoruyko2016wide}.
Our baseline model (on which most experiments are performed) is 28 layers deep with a width multiplier of 10, and is denoted by \wrn-28-10.
We also use deeper (up to 70 layers) and wider (up to 20) models.
Our largest model is a \wrn-70-16 containing 267M parameters, our smallest model is a \wrn-28-10 containing 36M parameters.
A popular option in the literature is the \wrn-34-20, which contains 186M parameters.

\vspace{-.2cm}
\paragraph{Outer minimization.}

We use \gls*{sgd} with Nesterov momentum~\citep{polyak1964some,nesterov27method}.
The initial learning rate of 0.1 is decayed by $10\times$ half-way and three-quarters-of-the-way through training (we refer to this schedule as the \emph{multistep} schedule).
We use a global weight decay parameter of $5\times10^{-4}$.
In the basic setting without additional unlabeled data, we use a batch size of $128$ and train for $200$ epochs (i.e., 78K steps).
With additional unlabeled data, we use a batch size of $1024$ with $512$ samples from \cifar and $512$ samples from a subset of 500K images extracted from the tiny images dataset \tinyimages~\citep{80m}~\footnote{We use the dataset from \citet{carmon_unlabeled_2019} available at \url{https://github.com/yaircarmon/semisup-adv}.} and train for 19.5K steps (i.e., $400$ \cifar-equivalent epochs).
To use the unlabeled data with adversarial training, we use the \emph{pseudo-labeling} mechanism presented by \cite{carmon_unlabeled_2019}: a separate non-robust classifier is trained on clean data from \cifar to provide labels to the unlabeled samples (the dataset created by \citealp{carmon_unlabeled_2019} is already annotated with such labels).
Our batches are split over $32$ Google Cloud TPU v$3$ cores.
As is common on \cifar, we augment our samples with random crops (i.e., pad by 4 pixels and crop back to $32\times32$) and random horizontal flips.

\vspace{-.2cm}
\paragraph{Inner maximization.}

The inner maximization in \autoref{eq:at_loss} is implemented using \pgd{10}.\footnote{Unless special care is taken \citep{wong_fast_2020}, using less than 7 steps often results in gradient obfuscated models \citep{qin_adversarial_2019}, using more steps did not provide significant improvements in our experiments}
We used a step-size $\alpha$ of $2/255$ and $15/255$ for \linf and \ltwo norm-bounded perturbations, respectively.
With this setup, training without additional data takes approximately $1$ hour and $30$ minutes for a \wrn-28-10 (peak accuracy as measured on a validation set disjoint from the test set is obtained after 1 hour and 15 minutes; for a \wrn-70-16 peak accuracy is obtained after 4 hours).
With additional unlabeled data, training takes approximately 2 hours.

\vspace{-.2cm}
\paragraph{Evaluation protocol.}

As adversarial training is a bit more noisy than regular training, for each hyperparameter setting, we train two models.
Throughout training we measure the robust accuracy using \pgd{40} on 1024 samples from a separate validation set (disjoint from the training and test set).
Similarly to \citet{rice_overfitting_2020}, we perform early stopping by keeping track of model parameters that achieve the highest robust accuracy (i.e., lowest adversarial risk as shown in \autoref{eq:adversarial_risk}) on the validation set.
From both models, we pick the one with highest robust accuracy on the validation set and use this model for a further more thorough evaluation.
Finally, the robust accuracy is reported on the full test set against a mixture of \autoattack~\citep{croce_reliable_2020} and \multitargeted~\citep{gowal_alternative_2019}.
We execute the following sequence of attacks: \autopgd on the cross-entropy loss with 5 restarts and 100 steps, \autopgd on the difference of logits ratio loss with 5 restarts and 100 steps, \multitargeted on the margin loss with 10 restarts and 200 steps.\footnote{For reference, the \autoattack leaderboard at \url{https://github.com/fra31/auto-attack} evaluates the model from \citet{rice_overfitting_2020} to 53.42\%, while this evaluation computes a robust accuracy of 53.38\% for the same model. The \autoattack leaderboard also evaluates the model from \citet{carmon_unlabeled_2019} to 59.53\%, while this evaluation computes a robust accuracy of 59.47\% for the same model.}
We also report the clean accuracy which is the top-1 accuracy without adversarial perturbations.

\subsection{Baseline.}
\label{sec:baseline}

As a comparison point, we train a \wrn-28-10 ten times with the default settings given above for both the \cifar-only and the additional data settings.
The resulting robust accuracy on the test set is 50.80$\pm$0.23\% (for \cifar-only) and 58.41$\pm$0.25\% (with additional unlabeled data).
When using a \wrn-34-20, we obtain 52.91\% which is in line with the model obtained by \citet{rice_overfitting_2020} for the same settings (i.e., 53.38\%).
When using TRADES (instead of regular adversarial training) in the additional data setting, we obtain 59.45\% which is in line with the model obtain by \citet{carmon_unlabeled_2019} for the same settings (i.e., 59.47\%).
As our pipeline is implemented in \texttt{JAX}~\citep{bradbury_jax_2018} and \texttt{Haiku}~\citep{hennigan_haiku_2020}, we do not exclude some slight differences in data preprocessing and network initialization.

\section{Experiments and analysis}
\label{sec:experiments}

The following sections detail different independent experiments (more experiments are available in the appendix).
Each section is self-contained to allow the reader to jump to any section of interest.
The outline is as follows:
\vspace{\parskip}
{
\setlength{\parskip}{0em}
\etocsettocstyle{}{}
\localtableofcontents
}

\subsection{Losses for Outer minimization}
\label{sec:outer_minimization}

As explained in \autoref{sec:at}, adversarial training as proposed by~\citet{madry_towards_2017} aims to minimize the loss given in \autoref{eq:at_loss} and is usually implemented as
\begin{align}
\mathcal{L}_\vtheta^\textrm{AT} = \xent(f(\vx + \hat{\vdelta}; \vtheta), y), ~~\mbox{ where }
\hat{\vdelta} \approx \argmax_{\vdelta \in \sS} \xent(f(\vx + \vdelta; \vtheta), y).
\end{align}
Here, $\xent$ denotes the cross-entropy loss and $\hat{\vdelta}$ is treated as a constant (i.e., there is no back-propagation through the inner optimization procedure). 
One of most successful variant of adversarial training is TRADES~\citep{zhang_theoretically_2019} which derives a theoretically grounded regularizer that balances the trade-off between standard and robust accuracy.
The overall loss used by TRADES is given by
\begin{equation}
\mathcal{L}_\vtheta^\textrm{TRADES} = \xent(f(\vx; \vtheta), y) + \beta \maximize_{\vdelta \in \sS} \kl(f(\vx + \vdelta; \vtheta), f(\vx; \vtheta)) \label{eq:trades_loss},
\end{equation}
where $\kl$ denotes the Kullback-Leibler divergence.
TRADES is one of the core components used by \citet{carmon_unlabeled_2019} within RST and by \citet{uesato_adversarial_2018} within UAT-OT.
A follow-up work \citep{wang_improving_2020}, known as Misclassification Aware Adversarial Training (MART), introduced a boosted loss that differentiates between the misclassified and correctly classified examples in a bid to improve this trade-off further.
We invite the reader to refer to~\citet{wang_improving_2020} for more details.

\begin{table}[t]
\begin{center}
\caption{\footnotesize{Clean (without perturbations) and robust (under adversarial attack) accuracy obtained by Adversarial Training (AT), TRADES and MART  trained on \cifar (with and without additional unlabeled data) against \linf norm-bounded perturbations of size $\epsilon = 8/255$.}
\label{table:at_trades_mart_main}}
\resizebox{.95\textwidth}{!}{
\begin{tabular}{p{0.42\textwidth}|cc|cc}
    \hline
    \cellcolor{header} & \multicolumn{2}{c|}{\cellcolor{header} \cifar} & \multicolumn{2}{c}{\cellcolor{header} with \tinyimages} \Tstrut \\
    \cellcolor{header} \textsc{Setup} & \cellcolor{header} \textsc{Clean} & \cellcolor{header} \textsc{Robust} & \cellcolor{header} \textsc{Clean} & \cellcolor{header} \textsc{Robust} \Bstrut \\
    \hline
    AT \citep{madry_towards_2017} & 84.85$\pm$1.20\% & 50.80$\pm$0.23\% & 90.93$\pm$0.25\% & 58.41$\pm$0.25\% \Tstrut \\
    TRADES \citep{zhang_theoretically_2019} & 82.74\% & 51.91\% & 88.36\% & 59.45\% \\
    MART \citep{wang_improving_2020} & 80.51\% & 51.93\% & 90.45\% & 58.25\% \Bstrut \\
    \hline
\end{tabular}
}
\end{center}
\end{table}

\paragraph{Results.}
\autoref{table:at_trades_mart_main} shows the performance of TRADES and MART compared to adversarial training (AT).
We observe that while TRADES systematically improves upon AT in both data settings, this is not the case for MART.
We find that MART has the propensity to create moderate forms of gradient masking against weak attacks like \pgd{20}.
In one extreme case, a \wrn-70-16 trained with additional unlabeled data using MART obtained an accuracy of 71.08\% against \pgd{20} which dropped to 60.63\% against our stronger suite of attacks (this finding is consistent with the AutoAttack leaderboard at \url{https://github.com/fra31/auto-attack}).

\paragraph{Key takeaways.}
Contrary to the suggestion of~\citet{rice_overfitting_2020} (i.e., \emph{``the original PGD-based adversarial training method can actually achieve the same robust performance as state-of-the-art method''}, see \autoref{sec:related_work}), TRADES (when combined with early-stopping -- as our setup dictates) is more competitive than classical adversarial training.
The results also highlight the importance of strong evaluations beyond \pgd{20} (including evaluations of the validation set used for early stopping).

\subsection{Inner maximization}
\label{sec:inner_maximization}

\subsubsection{Inner maximization loss}

Most works use the same loss (e.g., cross-entropy loss or Kullback-Leibler divergence) for solving the inner maximization problem (i.e., finding an adversarial example) and the outer minimization problem (i.e., training the neural network).
However, there are many plausible approximations for the inner maximization.
For example, instead of using the cross-entropy loss or Kullback-Leibler divergence for the inner maximization, \citet{uesato_adversarial_2018} used the margin loss $\max_{i \neq y} f(\vx; \vtheta)_i - f(\vx; \vtheta)_y$ which improved attack convergence speed.
Similarly, we could mix the cross-entropy loss with TRADES via the following:
\begin{align}
\mathcal{L}_\vtheta^\textrm{TRADES-XENT} &= \xent(f(\vx; \vtheta), y) + \beta \kl(f(\vx + \hat{\vdelta}; \vtheta), f(\vx; \vtheta)) \mbox{ where } \label{eq:trades_xent_loss} \\
\hat{\vdelta} &\approx \argmax_{\vdelta \in \sS} \xent(f(\vx + \vdelta; \vtheta), y). \nonumber
\end{align}
Or mix the Kullback-Leibler divergence with classical adversarial training:
\begin{align}
\mathcal{L}_\vtheta^\textrm{AT-KL} &= \xent(f(\vx + \hat{\vdelta}; \vtheta), y) \mbox{ where } \hat{\vdelta} \approx \argmax_{\vdelta \in \sS} \kl(f(\vx + \vdelta; \vtheta), f(\vx; \vtheta)).
\end{align}
Below are our observations on the effects of robustness and clean accuracy when we combine different inner and outer optimisation losses.

\begin{table}[t]
\begin{center}
\caption{\footnotesize{Clean (without perturbations) and robust (under adversarial attack) accuracy obtained by TRADES and Adversarial Training (AT) with different inner losses trained on \cifar (with and without additional unlabeled data) against \linf norm-bounded perturbations of size $\epsilon = 8/255$. The robust accuracy is measured using \pgd{40} with a margin loss (\textsc{Pgd}$^{40}_{\textrm{margin}}$) and our combination of \autoattack and \multitargeted (AA+MT).}
\label{table:inner_max_loss}}
\resizebox{.95\textwidth}{!}{
\begin{tabular}{l|ccc|ccc}
    \hline
    \cellcolor{header} & \multicolumn{3}{c|}{\cellcolor{header} \cifar} & \multicolumn{3}{c}{\cellcolor{header} with \tinyimages} \Tstrut \\
    \cellcolor{header} \textsc{Setup} & \cellcolor{header} \textsc{Clean} & \cellcolor{header} \textsc{Pgd}$^{40}_{\textrm{margin}}$ & \cellcolor{header} \textsc{AA+MT} & \cellcolor{header} \textsc{Clean} & \cellcolor{header} \textsc{Pgd}$^{40}_{\textrm{margin}}$ & \cellcolor{header} \textsc{AA+MT} \Bstrut \\
    \hline
    AT-XENT \citep{madry_towards_2017} & 84.85\% & 53.87\% & 50.80\% & 90.93\% & 61.46\% & 58.41\% \Tstrut \\
    AT-KL & 88.21\% & 50.67\% & 48.53\% & 91.86\% & 59.49\% & 56.89\% \\
    AT-MARGIN & 85.12\% & \textbf{54.72\%} & 48.79\% & 90.01\% & \textbf{61.53\%} & 55.18\% \\
    TRADES-XENT & 83.01\% & 54.19\% & \textbf{52.76\%} & 89.12\% & 61.25\% & 58.98\% \\
    TRADES-KL \citep{zhang_theoretically_2019} & 82.74\% & 53.85\% & 51.91\% & 88.36\% & 61.11\% & \textbf{59.45\%} \\
    TRADES-MARGIN & 81.60\% & 54.47\% & 51.28\% & 86.88\% & 61.37\% & 57.82\% \Bstrut \\
    \hline
\end{tabular}
}
\end{center}
\end{table}

\paragraph{Results.}
\autoref{table:inner_max_loss} shows the performance of different inner and outer loss combinations.
Each combination is evaluated against two adversaries: a weaker \pgd{40} attack using the margin loss (and denoted by \textsc{Pgd}$^{40}_{\textrm{margin}}$) and our stronger combination of attacks (as detailed in \autoref{sec:setup} and denoted AA+MT).
The first observation is that TRADES obtains higher robust accuracy compared to classical adversarial training (AT) across most combinations of inner losses.
We note that the clean accuracy is higher for AT compared to TRADES and address this trade-off in the next section (\autoref{sec:inner_radius}).
In the low-data setting, the combination of TRADES with cross-entropy (TRADES-XENT) yields the best robust accuracy.
In the high-data setting, we obtain higher robust accuracy using TRADES-KL.

Our second observation is that using margin loss during training can show signs of gradient masking: models trained using margin loss show higher degradation in robust accuracy when evaluated against the stronger adversary (AA+MT) as opposed to the weaker \textsc{Pgd}$^{40}_{\textrm{margin}}$.
\autoref{table:inner_max_loss} shows that the drop in robust accuracy can be as high as -6.35\%.\footnote{Using margin loss for evaluation is not uncommon, since it has been suggested to yield a stronger adversary (see \citealp{carlini_towards_2017,liu2016delving}).}
This level of gradient masking is most prominent in AT-MARGIN and is mitigated when we use TRADES for the outer minimization.
This degradation in robust accuracy is significantly reduced when we use cross-entropy as the inner maximization loss. 

\paragraph{Key takeaways.}
Similarly to the observation made in \autoref{sec:outer_minimization}, TRADES obtains higher adversarial accuracy compared to classical adversarial training. We found that using margin loss during training (for the inner maximization procedure) creates noticeable gradient masking.

\subsubsection{Inner maximization perturbation radius}
\label{sec:inner_radius}

Several works explored the use of larger~\citep{gowal_effectiveness_2018} or adaptive~\citep{balaji_instance_2019} perturbation radii.
Like TRADES, using different perturbation radii is an attempt to bias the clean to robust accuracy trade-off: as we increase the training perturbation radius we expect increased robustness and lower clean accuracy.

\paragraph{Results.}
\autoref{table:at_epsilon} shows the effect of increasing the perturbation radius $\epsilon$ by a factor $1.1\times$ and $1.2\times$ the original value of $8/255$ (during training, not during evaluation).
We notice that adversarial training (AT) can close the gap to TRADES as we increase the perturbation radius (especially in the low-data regime).
Although not reported here, we noticed that TRADES does not get similar improvements in performance with larger radii (possibly because TRADES is already actively managing the trade-off with clean accuracy).

\begin{table}[t]
\begin{center}
\caption{\footnotesize{Clean (without perturbations) and robust (under adversarial attack) accuracy obtained on \cifar (with and without additional unlabeled data) against \linf norm-bounded perturbations of size $\epsilon = 8/255$ when using different perturbation radii for the inner maximization trained.}
\label{table:at_epsilon}}
\resizebox{.95\textwidth}{!}{
\begin{tabular}{p{0.42\textwidth}|cc|cc}
    \hline
    \cellcolor{header} & \multicolumn{2}{c|}{\cellcolor{header} \cifar} & \multicolumn{2}{c}{\cellcolor{header} with \tinyimages} \Tstrut \\
    \cellcolor{header} \textsc{Setup} & \cellcolor{header} \textsc{Clean} & \cellcolor{header} \textsc{Robust} & \cellcolor{header} \textsc{Clean} & \cellcolor{header} \textsc{Robust} \Bstrut \\
    \hline
    AT $\epsilon = 8/255$ \citep{madry_towards_2017} & 84.85$\pm$1.20\% & 50.80$\pm$0.23\% & 90.93$\pm$0.25\% & 58.41$\pm$0.25\% \Tstrut \\
    AT $\epsilon = 8.8/255$ & 82.13\% & 51.65\% & 90.16\% & 58.72\% \\
    AT $\epsilon = 9.6/255$ & 83.60\% & 51.81\% & 89.39\% & 58.77\% \\
    TRADES $\epsilon = 8/255$ \citep{zhang_theoretically_2019} & 82.74\% & 51.91\% & 88.36\% & 59.45\% \Bstrut \\
    \hline
\end{tabular}
}
\end{center}
\end{table}

\paragraph{Key takeaways.}
Tuning the training perturbation radius can marginally improve robustness (when using classical adversarial training). 
We posit that understanding when and how to use larger perturbation radii might be critical towards our understanding of robust generalization.
Work from \citet{balaji_instance_2019} do provide additional insights, but the topic remains largely under-explored.

\subsection{Additional unlabeled data}
\label{sec:unlabeled}

After the work from \citet{schmidt_adversarially_2018} which posits that robust generalization requires more data, \citet{hendrycks_using_2019} demonstrated that one could leverage additional labeled data from \imagenet to improve the robust accuracy of models on \cifar.
\citet{uesato_are_2019} and \citet{carmon_unlabeled_2019} were among the first to introduce additional unlabeled data to \cifar by extracting images from \tinyimages (i.e., subset of high scoring images from a \cifar classifier).
\citet{uesato_are_2019} used 200K additional images to train their best model as they observed that using more data (i.e., 500K images) worsen their results.
This suggests that additional data needs to be close enough to the original \cifar images to be useful.
This is something already suggested by~\cite{oliver2018realistic} in the context of semi-supervised learning.

\subsubsection{Quality and quantity}

In this experiment, we use four different subsets of additional unlabeled images extracted from \tinyimages.
The first set with 500K images is the additional data used by \citet{carmon_unlabeled_2019}.\footnote{\url{https://github.com/yaircarmon/semisup-adv}}
The second, third and fourth sets consist of 200K, 500K and 1M images regenerated using a process identical to \citet{carmon_unlabeled_2019} with a another pre-trained classifier (achieving 95.86\% accuracy on the test set).
To generate a dataset of size $N$, we remove duplicates from the \cifar test set, score the remaining images using a standard \cifar classifier, and pick the top-$N/10$ scoring images from each class.
Hence, the dataset with 500K images contains all the images from the dataset with 200K images and, similarly, the dataset with 1M images contains all the images from the dataset with 500K images.
As the datasets increase in size, the images they contain may become less relevant to the \cifar classification task (as their standard classifier score becomes smaller).

%
\begin{wraptable}{r}{.45\textwidth}
\vspace{-1.5cm}
\caption{Accuracy under \linf attacks of size $\epsilon = 8/255$ on \cifar as the quantity of unlabeled data increases.\label{table:quantity_unlabeled}}
\vspace{-.3cm}
\begin{center}
\resizebox{.45\textwidth}{!}{
\begin{tabular}{l|cc}
    \hline
    \cellcolor{header} \textsc{Quantity} & \cellcolor{header} \textsc{Clean} & \cellcolor{header} \textsc{Robust} \TBstrut \\
    \hline
    500K \citep{carmon_unlabeled_2019} & 90.93\% & 58.41\% \Tstrut \\
    200K (regenerated) & 90.95\% & 57.29\% \\
    500K (regenerated) & 90.68\% & 59.12\% \\
    1M (regenerated) & 91.00\% & 58.89\% \Bstrut \\
    \hline
\end{tabular}
}
\end{center}
\vspace{-1.5cm}
\end{wraptable}
%

\paragraph{Results.}
\autoref{table:quantity_unlabeled} shows the performance of adversarial training with \emph{pseudo-labeling} as the quantity of additional unlabeled data increases.
This training scheme is identical to UAT-FT (as introduced by \citealp{uesato_are_2019}) and is slightly different to the one proposed in \citet{carmon_unlabeled_2019} (which used TRADES).
Firstly, we note that our regenerated set of 500K images improves robustness (+0.71\%) compared to \citet{carmon_unlabeled_2019}.
Secondly, similar to the observations made by \citet{uesato_are_2019}, there is a sweet spot where additional data is maximally useful.
Going from 200K to 500K additional images improves robust accuracy by +1.83\%.
However, increasing the amount of additional images further to 1M is detrimental (-0.23\%).
This suggests that more data improves robustness as long as the additional images relate to the original \cifar dataset (e.g., the more images we extract, the less likely it is that these images correspond to classes within \cifar).

\paragraph{Key takeaways.}
Small differences (e.g., different classifiers used for \emph{pseudo-labeling}) in the process that extracts additional unlabeled data can have significant impact on robustness (i.e., models trained on our regenerated dataset obtain higher robust accuracy than those trained with the data from \citealp{carmon_unlabeled_2019}).
There is also a trade-off between the quantity and the quality of the extra unlabeled data (i.e., increasing the amount of unlabeled data to 1M did not increase robustness).

\subsubsection{Ratio of labeled-to-unlabelled data per batch}
\label{sec:ratio_unlabeled}


In this section, all the experiments use the unlabeled data from \citet{carmon_unlabeled_2019}.
In the baseline setting, as done in \citet{carmon_unlabeled_2019}, each batch during training uses 50\% labeled data and the rest for unlabeled data.
This effectively downweighs unlabeled images by a factor 10$\times$, as we have 50K labeled images and 500K unlabeled images.
Increasing this ratio reduces the weight given to additional data and puts more emphasis of the original \cifar images.

%
\begin{wrapfigure}{r}{0.5\textwidth}
\begin{center}
\vspace{-1cm}
\includegraphics[width=.45\textwidth]{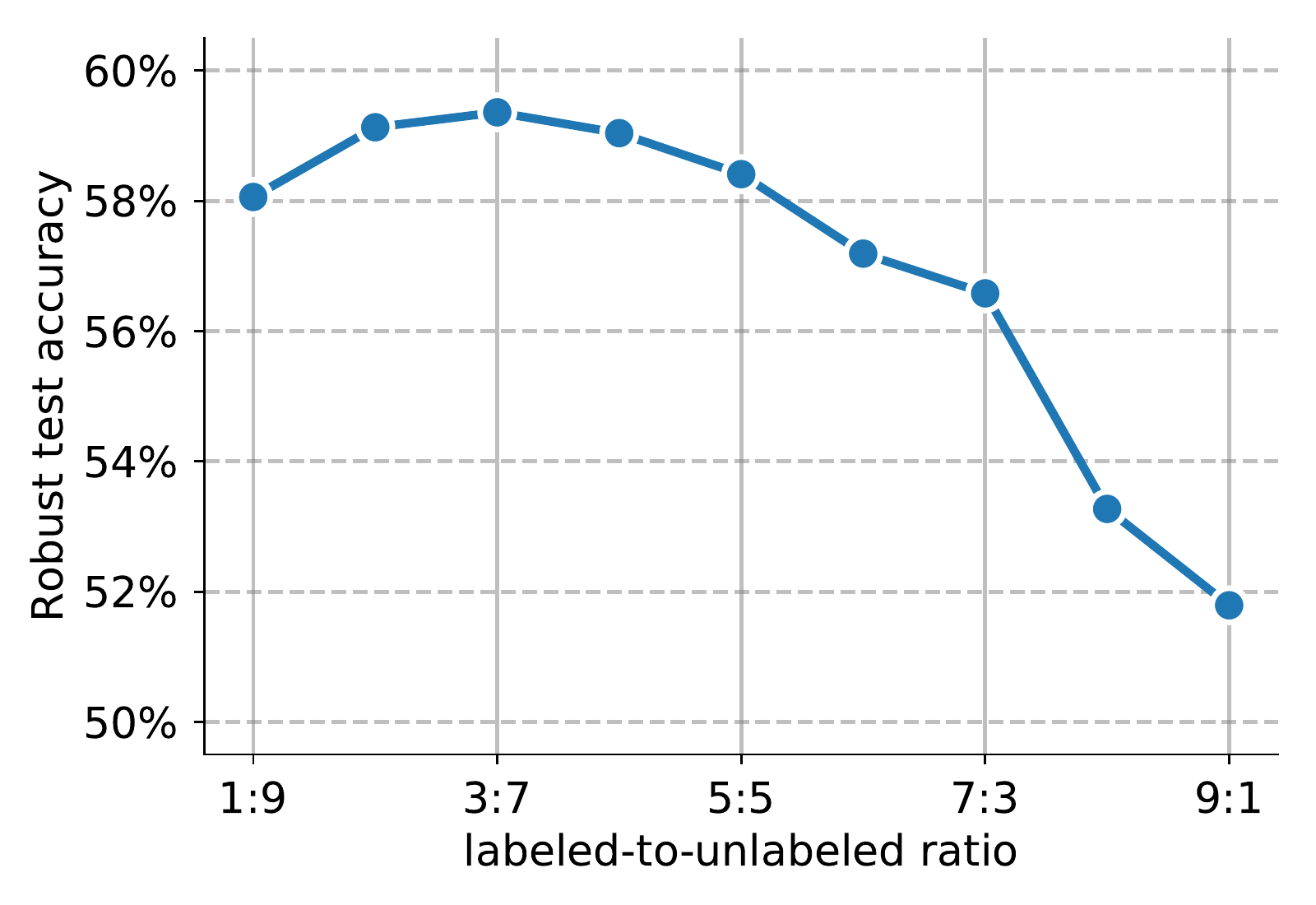}
\end{center}
\caption{Accuracy under \linf attacks of size $\epsilon = 8/255$ on \cifar as we vary the ratio of labeled-to-unlabeled data.\label{fig:ratio_summary_unlabeled_main}}
\vspace{-.3cm}
\end{wrapfigure}
%

\paragraph{Results.} 
\autoref{fig:ratio_summary_unlabeled_main} shows the robust accuracy as we vary that ratio.
We observe that giving slightly more importance to unlabeled data helps.
More concretely, we find an optimal ratio of labeled-to-unlabeled data of 3:7, which provides a boost of +0.95\% over the 1:1 ratio.
This suggest that the additional data extracted by \citet{carmon_unlabeled_2019} is well aligned with the original \cifar data and that we can improve robust generalization by allowing the model to see this additional data more frequently.
Increasing this ratio (i.e., reducing the importance of the unlabeled data) gradually degrades robustness and, eventually, the robust accuracy matches the one obtained by models trained without additional data.

We also experimented with label smoothing (for both labeled and unlabeled data independently).
Label smoothing should counteract the effect of the noisy labels resulting from the classifier used in the \emph{pseudo-labeling} process.
However, we did not observe improvements in performance for any of the settings tried.

\paragraph{Key takeaways.}
Tuning the weight given to unlabeled examples (by varying the labeled-to-unlabeled data ratio per batch) can provide improvements in robustness.
This experiment highlight that a careful treatment of the extra unlabeled data can provide improvements in adversarial robustness.

\subsection{Effects of scale}
\label{sec:modelsize}

\begin{figure*}[t]
\centering
\begin{subfigure}{0.49\textwidth}
\centering
\includegraphics[width=\linewidth]{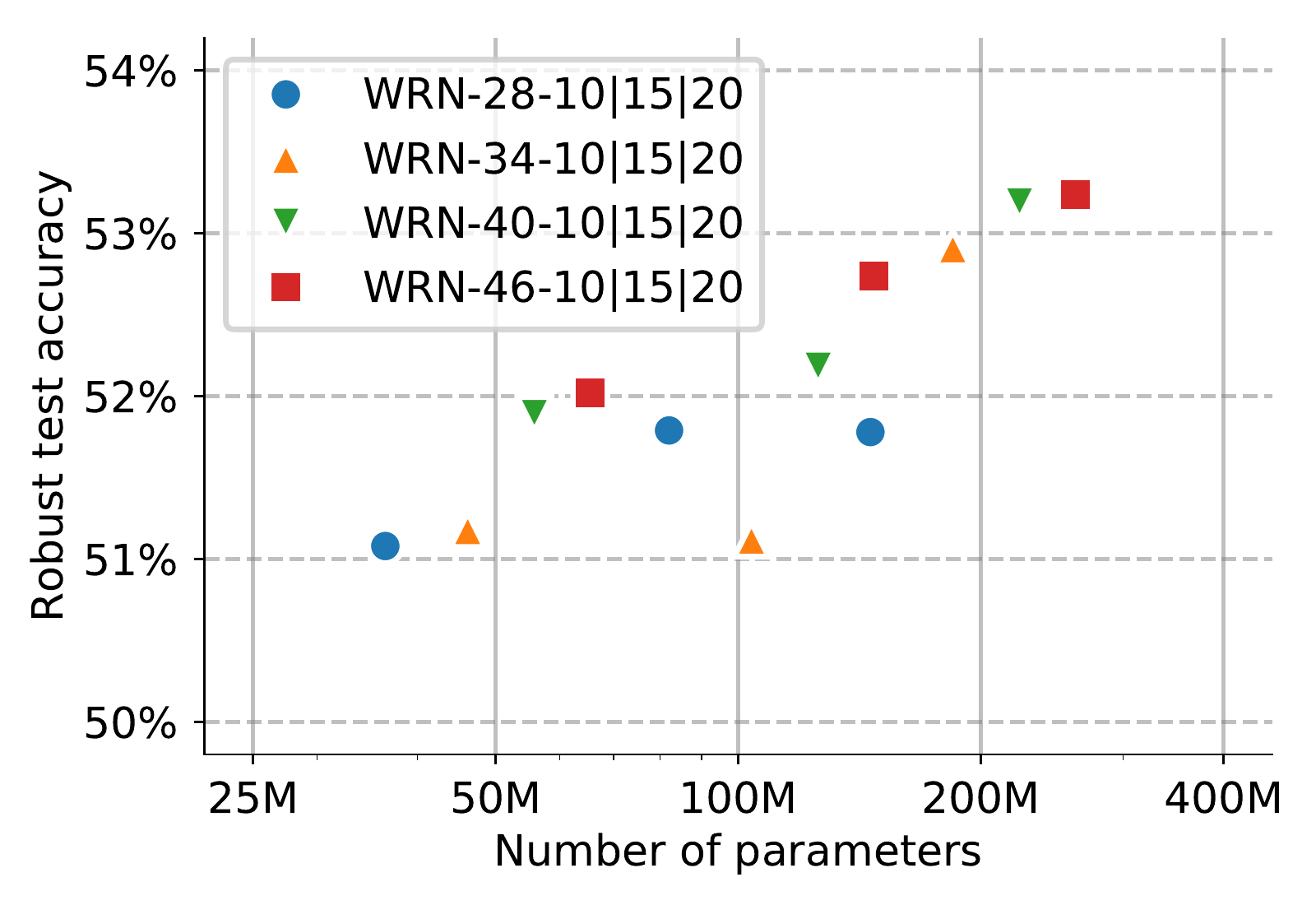}
\caption{\cifar\label{fig:width_depth_summary}}
\end{subfigure}
\begin{subfigure}{0.49\textwidth}
\centering
\includegraphics[width=\linewidth]{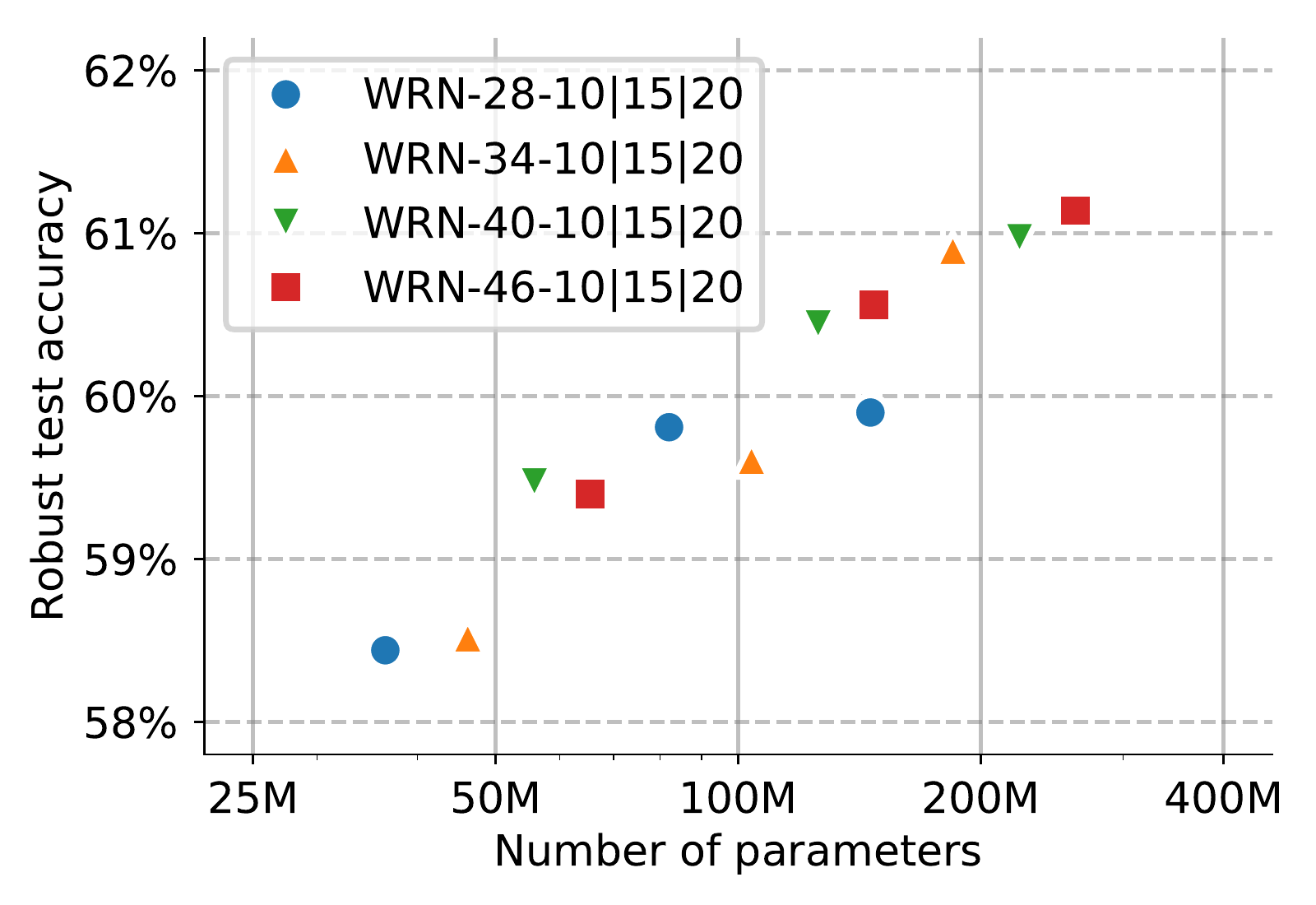}
\caption{\cifar and \tinyimages\label{fig:width_depth_summary_unlabeled}}
\end{subfigure}
\caption{Clean accuracy and accuracy under \linf attacks of size $\epsilon = 8/255$ on \cifar as the network architecture changes.
Panel \subref{fig:width_depth_summary} restricts the available data to \cifar, while panel \subref{fig:width_depth_summary_unlabeled} uses 500K additional unlabeled images extracted from \tinyimages.}
\label{fig:width_depth_summary_all}
\end{figure*}

\citet{rice_overfitting_2020} observed that increasing the model width improves robust accuracy despite the phenomenon of robust overfitting (which favors the use of early stopping in adversarial training).
\citet{uesato_are_2019} also trialed deeper models with a \wrn-106-8 and observed improved robustness.
A most systematic study of the effect of network depth was also conducted by \cite{xie2019intriguing} on \imagenet (scaling a ResNet to 638 layers).
However, there have not been any controlled experiments on \cifar that varied both depth and width of {\wrn}s.
We note that most work on adversarial robustness on \cifar use either a \wrn-34-10 or a \wrn-28-10 network, with \wrn-34-20 being another popular option.

\paragraph{Results.} 
\autoref{fig:width_depth_summary_all} shows the effect of increasing the depth and width of our baseline network.
It is possible to observe that, while both depth and width increase the number of effective model parameters, they do not always provide the same effect on robustness.
For example, a \wrn-46-15 which trains in roughly the same time as a \wrn-28-20 (about 5 hours in our setup) reaches higher robust accuracy: +0.96\% and +0.66\% on the settings without and with additional data, respectively.
\autoref{table:width_depth} in the appendix also shows that the clean accuracy improves as networks become larger.

\paragraph{Key takeaways.}
Larger models provide improved robustness~\citep{madry_towards_2017,xie_intriguing_2019,uesato_are_2019} and, for identical parameter count, deeper models can perform better.

\subsection{Other tricks}

\subsubsection{Model weight averaging}
\label{sec:ema}

Model Weight Averaging (WA)~\citep{izmailov_averaging_2018} is widely used in classical training~\citep{tan2019efficientnet}, and leads to better generalization.
The WA procedure finds much flatter solutions than \gls*{sgd}, and approximates ensembling with a single model.
To the best of our knowledge, the effect of WA on robustness have not been studied in the literature.
Ensembling has received some attention~\citep{pang2019improving,strauss2017ensemble}, but requires training multiple models.
We implement WA using an exponential moving average $\vtheta'$ of the model parameters $\vtheta$ with a decay rate $\tau$:
we execute $\vtheta' \gets \tau \cdot \vtheta' + (1 - \tau) \cdot \vtheta$ at each training step.
During evaluation, the weighted parameters $\vtheta'$ are used instead of the trained parameters $\vtheta$.

\begin{figure*}[t]
\centering
\begin{subfigure}{0.49\textwidth}
\centering
\includegraphics[width=\linewidth]{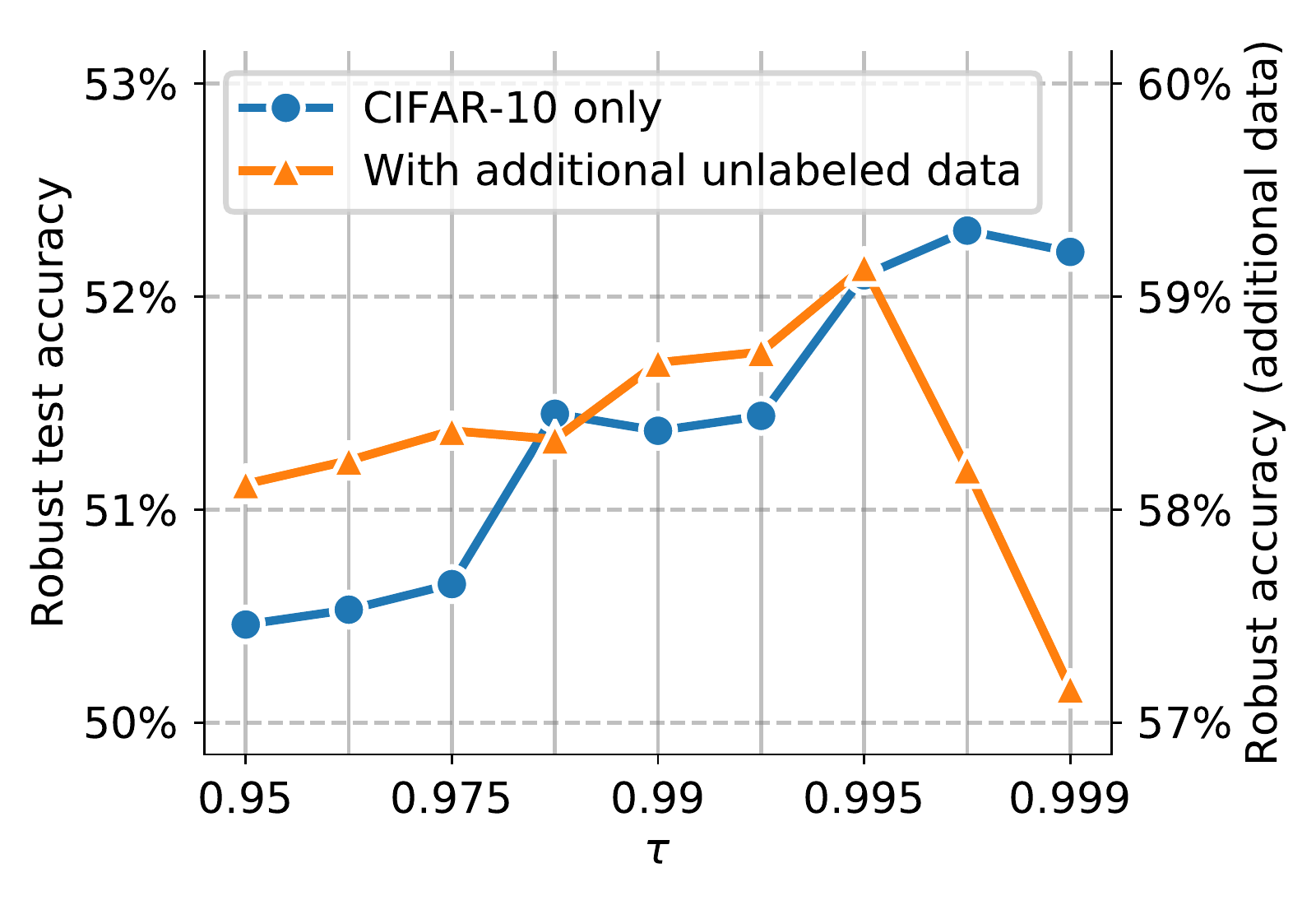}
\caption{Final robust accuracy\label{fig:ema_summary}}
\end{subfigure}
\begin{subfigure}{0.49\textwidth}
\centering
\includegraphics[width=\linewidth]{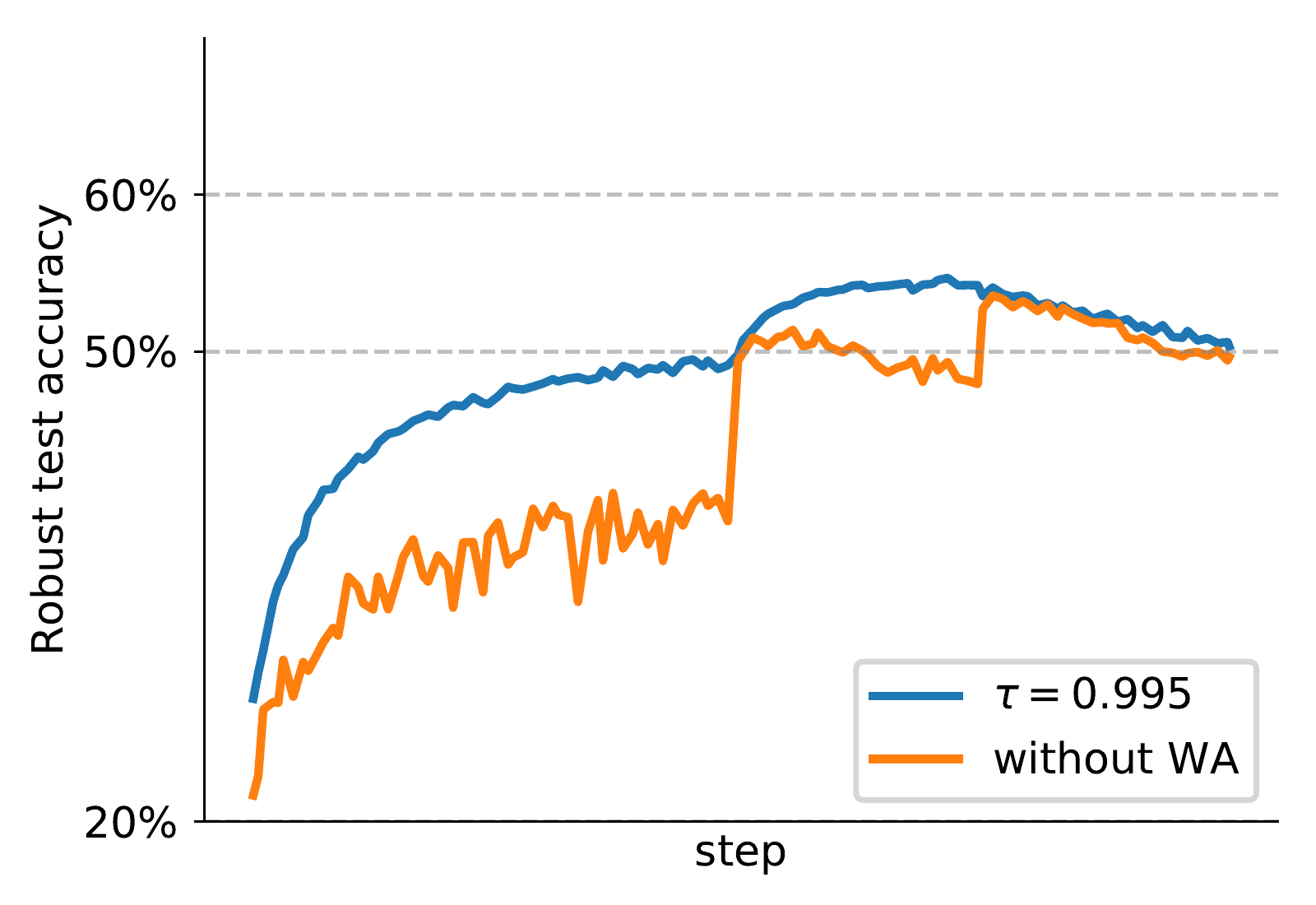}
\caption{Robust accuracy during training (\pgd{40})\label{fig:ema_comparison_main}}
\end{subfigure}
\caption{Accuracy under \linf attacks of size $\epsilon = 8/255$ on \cifar when using model weight averaging (WA).
Panel \subref{fig:ema_summary} shows the final robust accuracy obtained for different values of the decay rate $\tau$ for the settings without (blue and left y-axis) and with additional unlabeled data (orange and right y-axis).
Panel \subref{fig:ema_comparison_main} shows the evolution of the robust accuracy as training progresses.}
\label{fig:ema_main_all}
\end{figure*}

\paragraph{Results.}
\autoref{fig:ema_main_all} summarizes the performance of model weight averaging (detailed results are in \autoref{table:ema} in the appendix).
Panel~\ref{fig:ema_summary} demonstrates significant improvements in robustness: +1.41\% and +0.73\% with respect to the baseline without WA for settings without and with additional data, respectively.
In fact, in the low-data regime, WA provides an improvement similar to that of TRADES (+1.11\%).
A possible explanation for this phenomena is possibly given by panel~\ref{fig:ema_comparison_main}.
We observe that not only that WA achieves higher robust accuracy, but also that it maintains this higher accuracy over a few training epochs (about 25 epochs).
This reduces the sensitivity to early stopping which may miss the most robust checkpoint (happening shortly after the second learning rate decay).
Another important benefit of WA is its rapid convergence to about 50\% robust accuracy (against \pgd{40}) in the early stages of training, which suggests that it could be combined with efficient adversarial training techniques such as the one presented by~\citet{wong_fast_2020}.

\paragraph{Key takeaways.}
Although widely used for standard training, WA has not been explored within adversarial training.
We discover that WA provides sizeable improvements in robustness and hope that future work can explore this phenomenon.

\subsubsection{Activation functions}
\label{sec:activations}

With the exception from work by \citet{xie2020smooth} which focused on \imagenet, there have been little to no investigations into the effect of different activation functions on adversarial training.
\citet{xie2020smooth} discovered that ``smooth'' activation functions yielded higher robustness when using weak adversaries during training (PGD with a low number of steps).
In particular, they posit that they allow adversarial training to find harder adversarial examples and compute better gradient updates.
\citet{qin_adversarial_2019} also experimented with \texttt{softplus} activations with success.

\begin{figure*}[t]
\centering
\begin{subfigure}{0.49\textwidth}
\centering
\includegraphics[width=\linewidth]{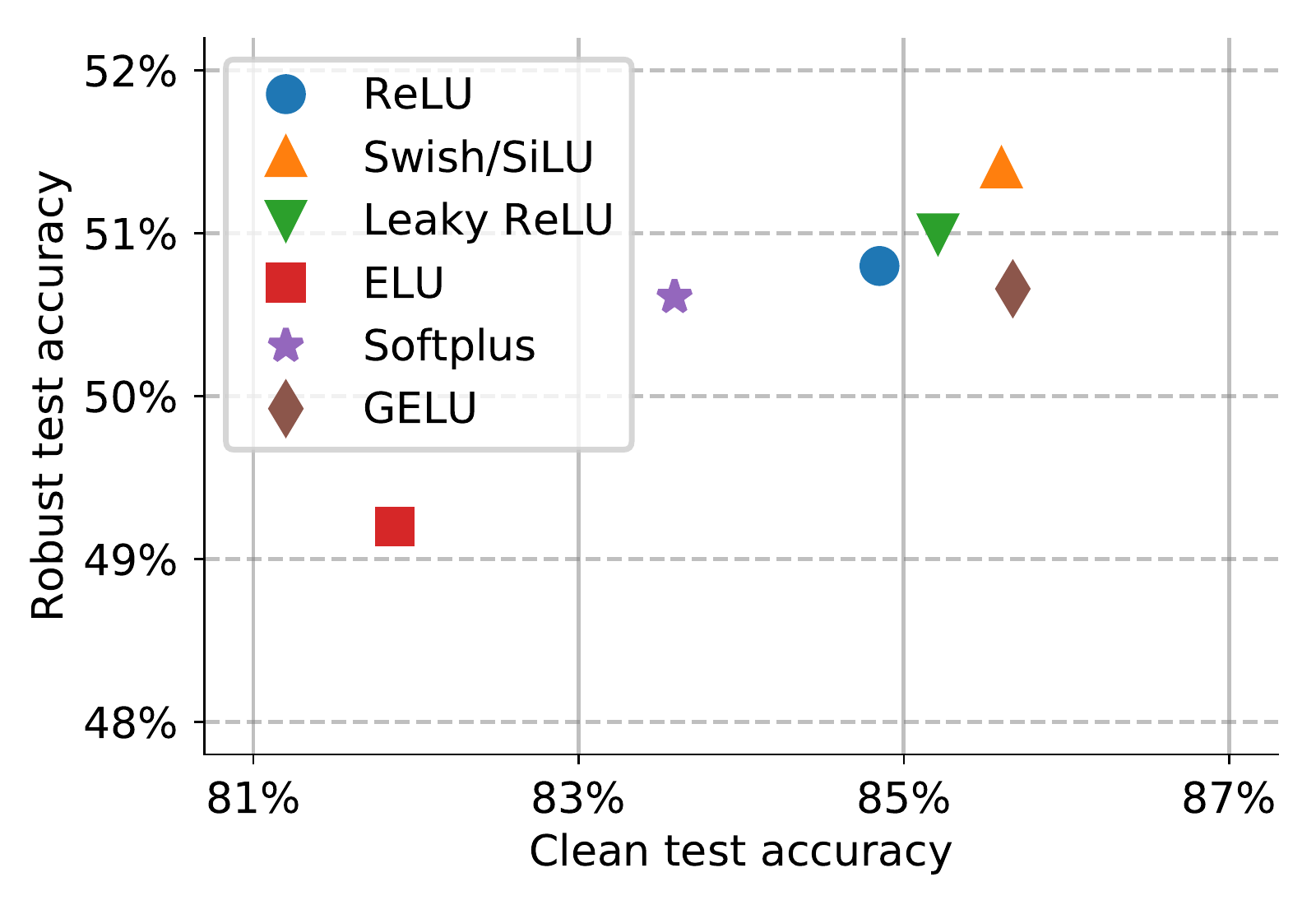}
\caption{\cifar\label{fig:activation_summary}}
\end{subfigure}
\begin{subfigure}{0.49\textwidth}
\centering
\includegraphics[width=\linewidth]{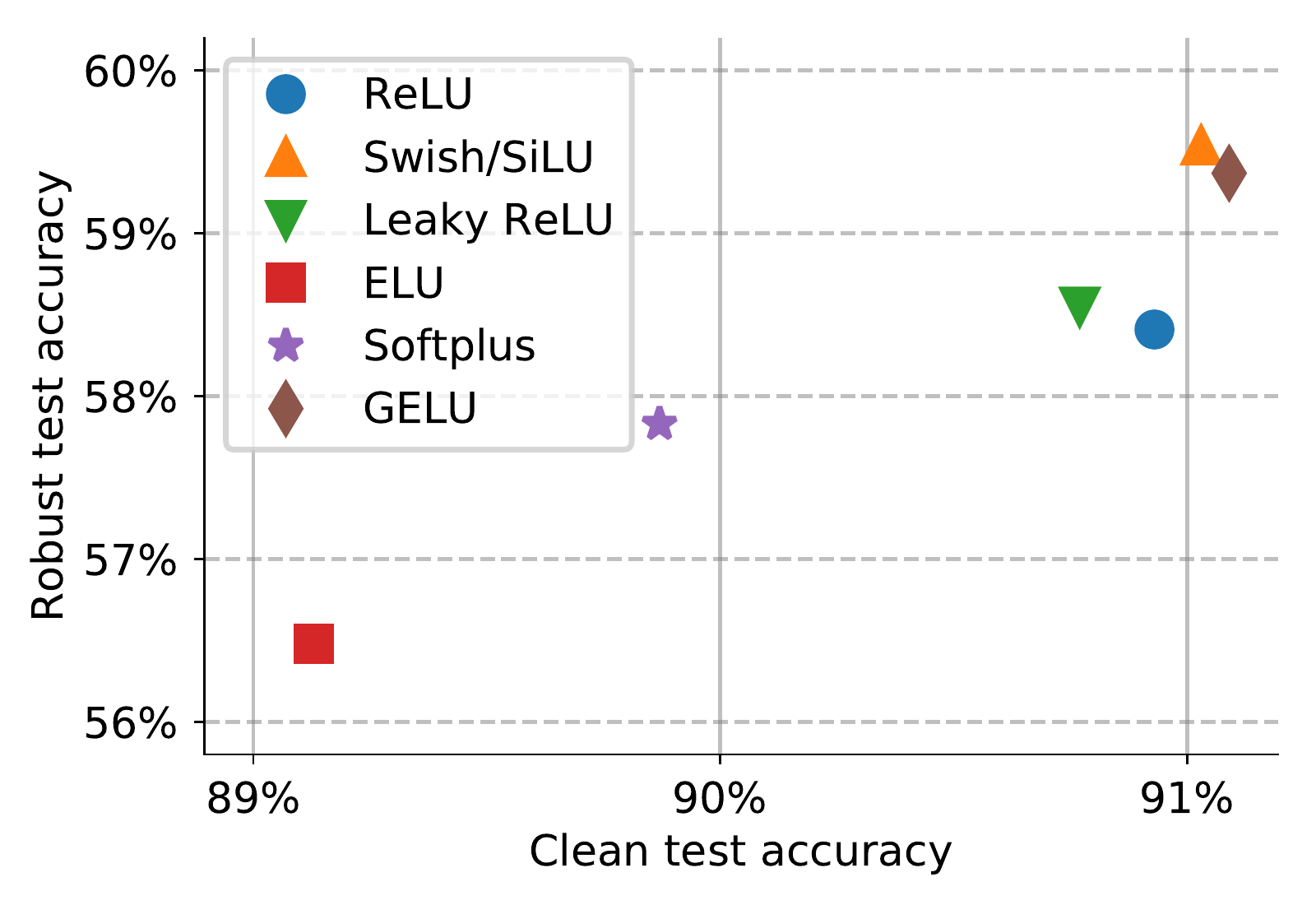}
\caption{\cifar and \tinyimages\label{fig:activation_summary_unlabled}}
\end{subfigure}
\caption{Clean accuracy and accuracy under \linf attacks of size $\epsilon = 8/255$ on \cifar for different activation functions.
Panel \subref{fig:activation_summary} restricts the available data to \cifar, while panel \subref{fig:activation_summary_unlabled} uses 500K additional unlabeled images extracted from \tinyimages.}
\label{fig:activation_summary_all}
\end{figure*}

\paragraph{Results.}
While we observe in \autoref{fig:activation_summary_all} that activation functions other than \texttt{ReLU}~\citep{nair2010rectified} can positively affect clean and robust accuracy, the trend is not as clear as the one observed by \citet{xie2020smooth} on \imagenet.
In fact, apart from \swish~\citep{hendrycks2016gaussian,ramachandran2017searching,elfwing2017sigmoidweighted}, which provides improvements of +0.8\% and +1.13\% in the settings without and with additional unlabeled data, the order of the best performing activation functions changes when we go from the \cifar-only setting to the setting with additional unlabeled data.
Overall, \texttt{ReLU} remains a good choice.

\paragraph{Key takeaways.}
The choice of activation function matters: we found that while \swish performs best in our experiments. Other ``smooth'' activation functions~\citep{xie_smooth_2020} do not necessarily correlate positively with robustness in our experiments.

\section{A new state-of-the-art}
\label{sec:combined}

To evaluate the combined effects of our findings, we train a single model according to the key takeaways from each section.
We hope that this exercise can provide valuable insights.
In particular, we combine the following elements: \textit{(i)} TRADES as detailed by \citet{zhang_theoretically_2019}, \textit{(ii)} model weight averaging with $\tau = 0.995$ (per $1024$ examples), \textit{(iii)} a larger model with a \wrn-70-16 (we also evaluate smaller models on \cifar against \linf norm-bounded perturbations), \textit{(iv)} \swish activations.
In the setting without additional labeled data, we use the \emph{multistep} learning rate schedule (as detailed in \autoref{sec:setup}) with a batch size of $512$ (the effective weight averaging decay $\tau$ is set to $\sqrt{0.995}$).
In the setting with additional labeled data, we preferred the \emph{cosine} decay learning rate schedule~\citep{SGDR} without restarts (which was better than \emph{multistep} when the model size increased beyond \wrn-28-10).
We also used a labeled-to-unlabeled ratio of 3:7.

\begin{table}[b]
\begin{center}
\caption{\footnotesize{Clean (without perturbations) and robust (under adversarial attack) accuracy obtained by different setups (with and without additional unlabeled data). The accuracies are reported on full test sets. * Trained by us using standard adversarial training with early stopping.}
\label{table:combined}}
\resizebox{1.\textwidth}{!}{
\begin{tabular}{l|ccc|c|ccc}
    \hline
    \cellcolor{header} \textsc{Model} & \cellcolor{header} \textsc{Dataset} & \cellcolor{header} \textsc{Norm} & \cellcolor{header} \textsc{Radius} &  \cellcolor{header} \textsc{Clean} & \cellcolor{header} \textsc{\pgd{40}} & \cellcolor{header} \textsc{AA+MT} & \cellcolor{header} \textsc{AA} \TBstrut \\
    \hline
    \hline
    \multicolumn{8}{l}{\textsc{\cellcolor{subheader} Without \tinyimages}} \TBstrut \\
    \hline
    \cite{pang_boosting_2020} (\wrn-34-20) & \multirow{3}{*}{\cifar} & \multirow{3}{*}{\linf} & \multirow{3}{*}{$\epsilon = 8/255$} & 85.14\% & -- & -- & 53.74\% \Tstrut \\
    Ours (\wrn-70-16) & & & & 85.29\% & 58.22\% & 57.14\% & \textbf{57.20\%} \\
    Ours (\wrn-34-20) & & & & 85.64\% & 57.73\% & 56.82\% & 56.86\% \Bstrut \\
    \hline
    \cite{robustness} (ResNet-50) & \multirow{2}{*}{\cifar} & \multirow{2}{*}{\ltwo} & \multirow{2}{*}{$\epsilon = 128/255$} & 90.83\% & -- & -- & 69.24\% \Tstrut \\
    Ours (\wrn-70-16) & & & & 90.90\% & 75.41\% & 74.45\% & \textbf{74.50\%} \Bstrut \\
    \hline
    \cite{rice_overfitting_2020} (ResNet-18) & \multirow{2}{*}{\cifarh} & \multirow{2}{*}{\linf} & \multirow{2}{*}{$\epsilon = 8/255$} & 53.83\% & -- & -- & 18.95\% \Tstrut \\
    Ours (\wrn-70-16) & & & & 60.86\% & 31.47\% & 30.67\% & \textbf{30.03\%} \Bstrut \\
    \hline
    \cite{zhang_towards_2019} (7-layer CNN) & \multirow{2}{*}{\mnist} & \multirow{2}{*}{\linf} & \multirow{2}{*}{$\epsilon = 0.3$} & 98.38\% & -- & -- & 93.96\% \Tstrut \\
    Ours (\wrn-28-10) & & & & 99.26\% & 97.27\% & 96.38\% & \textbf{96.34\%} \Bstrut \\
    \hline
    \cite{rice_overfitting_2020} (\wrn-28-10)* & \multirow{3}{*}{\svhn} & \multirow{3}{*}{\linf} & \multirow{3}{*}{$\epsilon = 8/255$} & 92.15\% & 57.82\% & 50.08\% & -- \Tstrut \\
    Ours (\wrn-34-20) & & & & 93.13\% & 61.12\% & \textbf{58.02\%} & -- \\
    Ours (\wrn-28-10) & & & & 92.87\% & 60.17\% & 56.83\% & -- \Bstrut \\
    \hline
    \hline
    \multicolumn{8}{l}{\textsc{\cellcolor{subheader} With Additional Unlabeled Data from \tinyimages}} \TBstrut \\
    \hline
    \cite{carmon_unlabeled_2019} (\wrn-28-10) & \multirow{4}{*}{\cifar} & \multirow{4}{*}{\linf} & \multirow{4}{*}{$\epsilon = 8/255$} & 89.69\% & -- & 59.47\% & 59.53\% \Tstrut \\
    Ours (\wrn-70-16) & & & & 91.10\% & 67.16\% & 65.87\% & \textbf{65.88\%} \\
    Ours (data from \citealp{carmon_unlabeled_2019}) & & & & 90.95\% & 66.70\% & 65.06\% & -- \\
    Ours (\wrn-28-10) & & & & 89.48\% & 64.08\% & 62.76\% & 62.80\% \Bstrut \\
    \hline
    \cite{augustin_adversarial_2020} (ResNet-50) & \multirow{2}{*}{\cifar} & \multirow{2}{*}{\ltwo} & \multirow{2}{*}{$\epsilon = 128/255$} & 91.08\% & -- & -- & 72.91\% \Tstrut \\
    Ours (\wrn-70-16) & & & & 94.74\% & 82.19\% & 80.45\% & \textbf{80.53\%} \\
    \hline
    \cite{hendrycks_using_2019} (ResNet-18) & \multirow{2}{*}{\cifarh} & \multirow{2}{*}{\linf} & \multirow{2}{*}{$\epsilon = 8/255$} & 59.23\% & -- & -- & 28.42\% \Tstrut \\
    Ours (\wrn-70-16) & & & & 69.15\% & 38.97\% & 37.70\% & \textbf{36.88\%} \Bstrut \\
    \hline
\end{tabular}
}
\end{center}
\end{table}

\autoref{table:combined} shows the results as well as the known state-of-the-art.
We evaluate robust accuracy using the \autoattack pipeline available from \url{https://github.com/fra31/auto-attack} and denoted AA, and using our combined set of attacks denoted AA+MT and described in \autoref{sec:setup}.
On \cifar against \linf norm-bounded perturbations of size $\epsilon = 8/255$, we improve state-of-the-art robust accuracy by +3.46\% and +6.35\% without and with additional data, respectively.
At equal model size, we improve robust accuracy by +3.12\% and +3.27\%, respectively.  
Most notably, without additional data, we improve on the results of four methods that used additional data~\citep{hendrycks_using_2019,uesato_are_2019,wang_improving_2020,sehwag2020hydra}.
With additional data, we surpass the barrier of 60\% for the first time.

To test the generality of our findings, we keep the same hyper-parameters and train adversarially robust models on \cifar against \ltwo norm-bounded perturbations of size $\epsilon = 128/255$ and on \cifarh against \linf norm-bounded perturbations of size $\epsilon = 8/255$.\footnote{For \cifarh, we extract new additional unlabeled data by excluding images from its test set.}
In the four cases (with and without additional data), we surpass the known state-of-the-art by significant margins.\footnote{For completeness, we also include results on \mnist and \svhn.}

\section{Conclusion}
In this work, we performed a systematic analysis of many different aspects surrounding adversarial training that can affect the robustness of trained networks.
The goal was to see how far we could push robust accuracy through adversarial training with the right combination of network size, activation functions, additional data and model weight averaging.
We find that by combining these different factors carefully we can achieve robust accuracy that improves upon the state-of-the-art by more than 6\% (in the setting using additional data on \cifar against \linf norm-bounded perturbations of size $\epsilon = 8/255$).
We hope that this work can serve as a reference point for the current state of adversarial robustness and can help others build new techniques that can ultimately reach higher adversarial robustness.



\bibliography{iclr2021_conference}
\bibliographystyle{iclr2021_conference}

\clearpage
\appendix

\section{Additional experiments}

In order to keep the main text concise, we relegated additional experiments to this section.
Similarly to \autoref{sec:experiments}, each section is self-contained to allow the reader to jump to any section of interest.
The outline is as follows:
\vspace{\parskip}
{
\setlength{\parskip}{0em}
\etocsettocstyle{}{}
\localtableofcontents
}

\subsection{Outer minimization}

\subsubsection{Learning rate schedule}

In this section, we test different learning rate schedules.
In particular, we compare the \emph{multistep} schedule introduced in \autoref{sec:setup}, where the initial learning rate of $0.1$ is decayed by $10\times$ half-way and three-quarters-of-the-way through training, with the \emph{cosine} and \emph{exponential} schedules.
For the \emph{cosine} schedule, we set the initial learning rate to $0.1$ and decay it to $0$ by the end of training.
For the \emph{exponential} schedule, we set the initial learning rate to $0.1$ and decay it every $5$ epochs such that by the end of training the final learning rate is $0.001$.
Learning rates are all scaled according to the batch size (i.e., $\textrm{effective learning} = \max(\textrm{learning rate} \times \textrm{batch size} / 256, \textrm{learning rate})$).

\begin{table}[b]
\begin{center}
\caption{\footnotesize{Clean (without perturbations) and robust (under adversarial attack) accuracy obtained by different learning rate schedules trained on \cifar (with and without additional unlabeled data) against \linf norm-bounded perturbations of size $\epsilon = 8/255$.}
\label{table:lr_schedule}}
\resizebox{.95\textwidth}{!}{
\begin{tabular}{p{0.42\textwidth}|cc|cc}
    \hline
    \cellcolor{header} & \multicolumn{2}{c|}{\cellcolor{header} \cifar} & \multicolumn{2}{c}{\cellcolor{header} with \tinyimages} \Tstrut \\
    \cellcolor{header} \textsc{Setup} & \cellcolor{header} \textsc{Clean} & \cellcolor{header} \textsc{Robust} & \cellcolor{header} \textsc{Clean} & \cellcolor{header} \textsc{Robust} \Bstrut \\
    \hline
    \hline
    \multicolumn{5}{l}{\cellcolor{header} \textsc{Learning Rate Schedule}} \TBstrut \\
    \hline
    Multistep decay & 84.85$\pm$1.20\% & 50.80$\pm$0.23\% & 90.93$\pm$0.25\% & 58.41$\pm$0.25\% \Tstrut \\
    Cosine & 83.90\% & 47.49\% & 91.28\% & 57.87\% \\
    Exponential & 83.39\% & 47.73\% & 91.08\% & 56.56\% \Bstrut \\
    \hline
\end{tabular}
}
\end{center}
\end{table}

\paragraph{Results.}

\autoref{table:lr_schedule} shows that, as they are currently implemented, the \emph{multistep} schedule is superior to the \emph{cosine} and \emph{exponential} schedules.
While, we did our best to tune all schedules, we do not exclude the possibility that better schedules exist.
In particular, the optimal schedule may depend on the model architecture and method used to find adversarial examples (e.g., AT or TRADES).
Smoother schedules (like \emph{cosine} or \emph{exponential}) are also less sensitive to the early stopping criterion and may result in less noisy results.
When using additional unlabeled data, model weight averaging (\autoref{sec:ema}) and a \wrn-70-16, we found that the \emph{cosine} schedule performed slightly better than the \emph{multistep} schedule (+0.24\%).

\paragraph{Key takeaways.}

The \emph{multistep} schedule developed over the years, which has been tuned to Wide-ResNets and adversarial training, works well for settings with and without additional unlabeled data.

\begin{table}[t]
\begin{center}
\caption{\footnotesize{Clean (without perturbations) and robust (under adversarial attack) accuracy obtained for different number of training epochs trained on \cifar (with and without additional unlabeled data) against \linf norm-bounded perturbations of size $\epsilon = 8/255$.}
\label{table:epochs}}
\resizebox{.95\textwidth}{!}{
\begin{tabular}{p{0.42\textwidth}|cc|cc}
    \hline
    \cellcolor{header} & \multicolumn{2}{c|}{\cellcolor{header} \cifar} & \multicolumn{2}{c}{\cellcolor{header} with \tinyimages} \Tstrut \\
    \cellcolor{header} \textsc{Setup} & \cellcolor{header} \textsc{Clean} & \cellcolor{header} \textsc{Robust} & \cellcolor{header} \textsc{Clean} & \cellcolor{header} \textsc{Robust} \Bstrut \\
    \hline
    \hline
    \multicolumn{5}{l}{\cellcolor{header} \textsc{Number of training epochs}} \TBstrut \\
    \hline
    50 & 78.83\% & 47.25\% & -- & -- \Tstrut \\
    100 & 84.30\% & 49.89\% & 85.62\% & 53.20\% \\
    200 & 84.85$\pm$1.20\% & 50.80$\pm$0.23\% & 89.74\% & 57.37\% \\
    400 & 83.11\% & 50.16\% & 90.93$\pm$0.25\% & 58.41$\pm$0.25\% \\
    800 & -- & -- & 91.09\% & 56.98\% \Bstrut \\
    \hline
\end{tabular}
}
\end{center}
\end{table}

\subsubsection{Number of optimization steps}

Training for longer is not always beneficial, especially when is comes to adversarial training.
The robust overfitting phenomenon studied by \citet{rice_overfitting_2020} attests to the difficulty of finding the right number of steps to optimize for.
In this experiment, we use the \emph{multistep} learning rate schedule and change the number of training epochs.
We expect to find an optimal schedule that balances overfitting with robustness.

\begin{wrapfigure}{r}{0.5\textwidth}
\begin{center}
\vspace{-1cm}
\includegraphics[width=.45\textwidth]{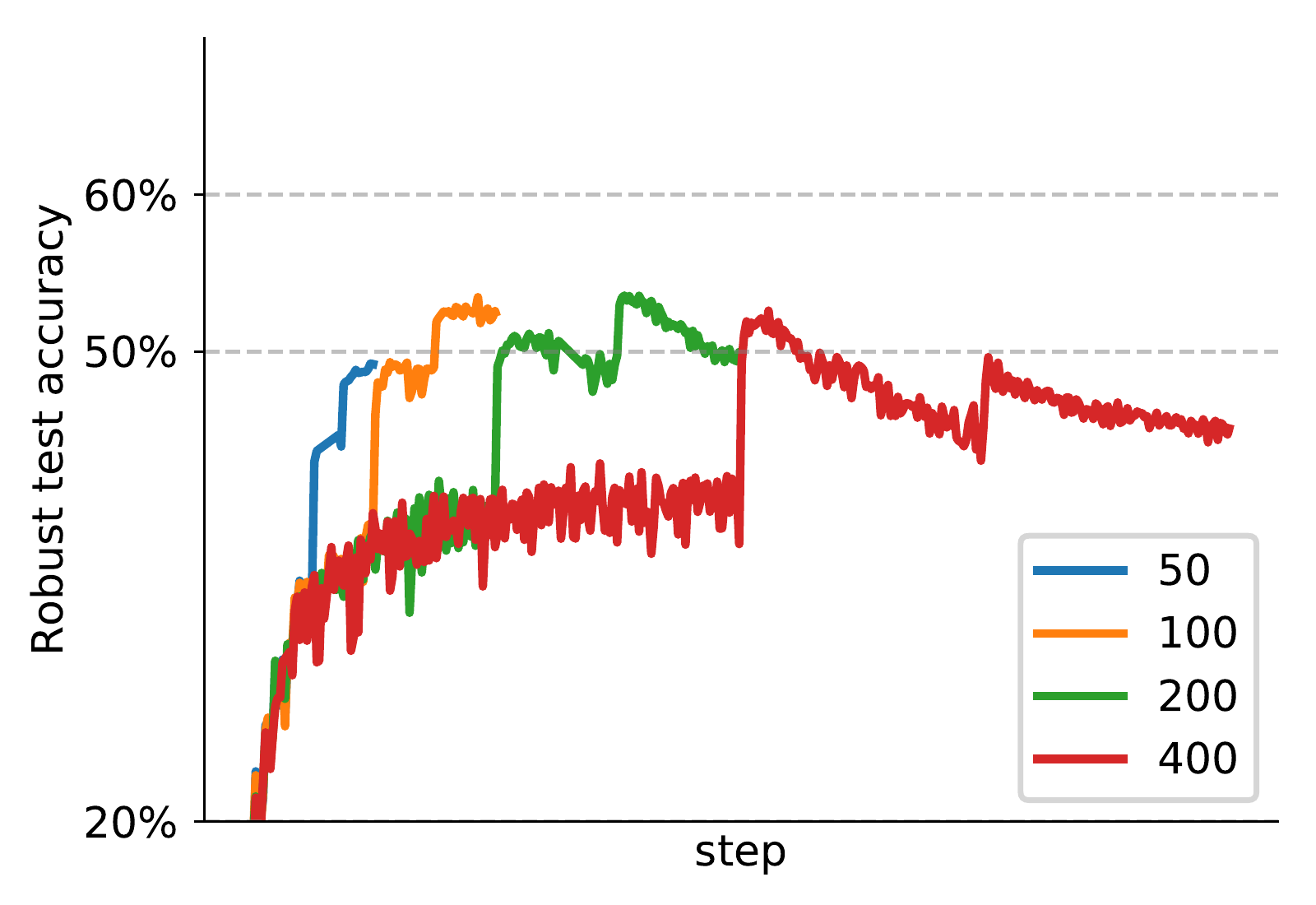}
\end{center}
\caption{Accuracy under \linf attacks of size $\epsilon = 8/255$ on \cifar as we vary the number of training epochs (without additional unlabeled data) against \pgd{40}.\label{fig:epochs}}
\vspace{-1cm}
\end{wrapfigure}

\paragraph{Results.}
In \autoref{table:epochs}, we vary the number of training epochs between $\{50, 100, 200, 400\}$ for the setting without additional data and between $\{100, 200, 400, 800\}$ for the setting with additional data.
For both settings, training for longer is not beneficial.
Without additional data, using 400 epochs instead of 200 leads to degradation of the robust accuracy by -0.64\%.
With additional data, using 800 epochs instead of 400 leads to degradation of -1.43\%.
\autoref{fig:epochs} shows the robust accuracy as training progresses for the setting without additional data.
We observe that letting the model train for longer leads to robust overfitting.

\paragraph{Key takeaways.}

Robust training (with adversarial training) does not benefit from longer training times.
The phenomenon of robust overfitting \citep{rice_overfitting_2020} can lead to reduced performance.
As such, it is important to balance the number of training epochs with other hyperparameters (such as \ltwo regularization).

\subsubsection{\ltwo regularization}

For completeness, we also explore explicit regularization using $\ell_2$ regularization on the model weights.
In this experiment, we vary the weight decay parameter between zero and $5\cdot10^{-3}$.

\begin{wrapfigure}{r}{0.5\textwidth}
\vspace{-1.15cm}
\begin{center}
\includegraphics[width=.45\textwidth]{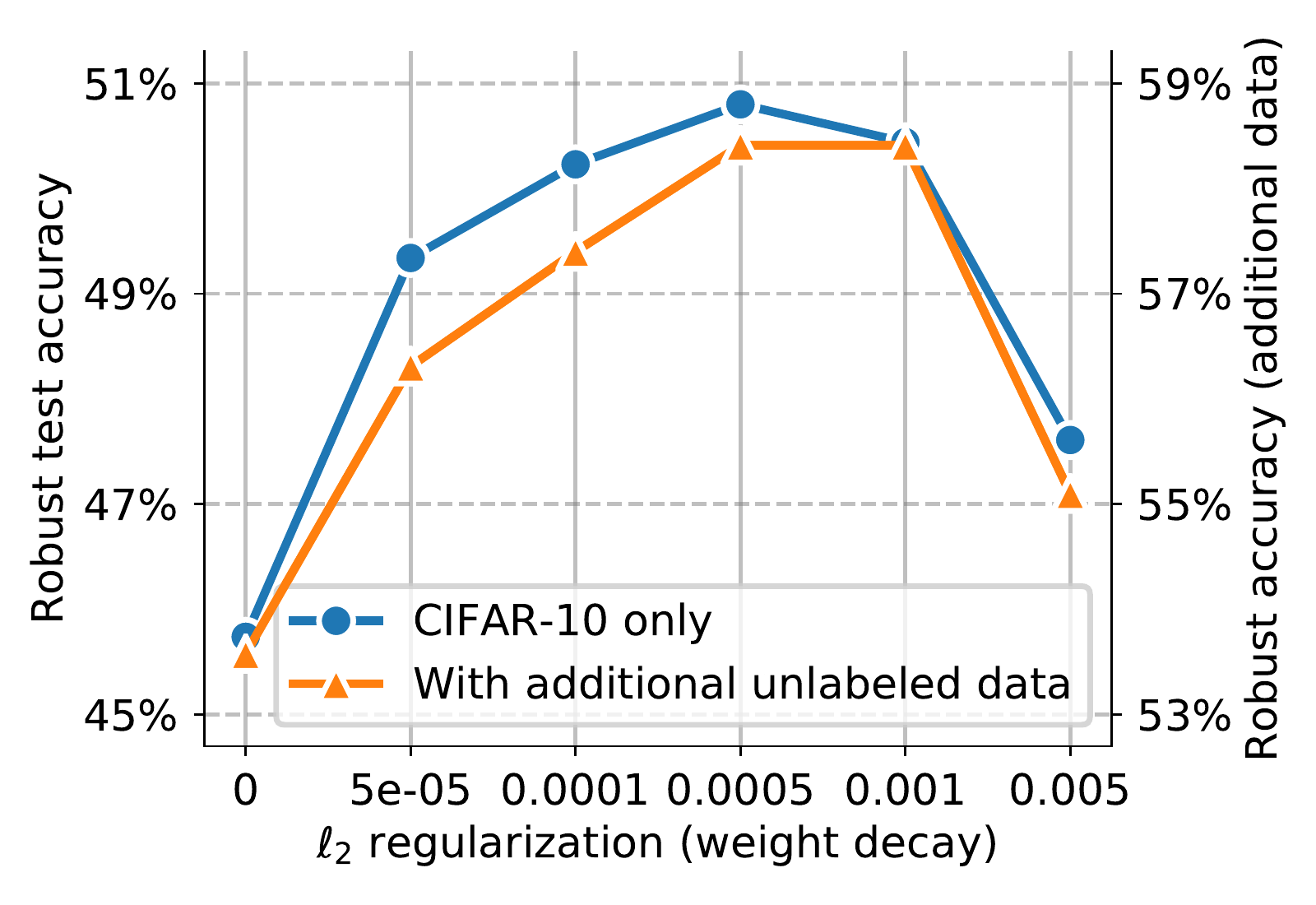}
\end{center}
\caption{Accuracy under \linf attacks of size $\epsilon = 8/255$ on \cifar as we vary weight decay.\label{fig:l2_reg}}
\vspace{-1cm}
\end{wrapfigure}

\paragraph{Results.}

\autoref{fig:l2_reg} (and \autoref{table:l2_reg} in \autoref{sec:additional_tables}) demonstrates that there exists an ideal weight decay value.
In particular, the best decay of $5\cdot10^{-4}$ works well across both data settings (with and without additional data).
This value is also the value used frequently in the literature when training adversarially robust Wide-ResNets \citep{carmon_unlabeled_2019,uesato_are_2019,rice_overfitting_2020,wong_fast_2020}.
Without any $\ell_2$ regularization, the robust accuracy drops by -5.06\% (in the setting without additional data).

\paragraph{Key takeaways.}

$\ell_2$ regularization is an important element of adversarial training.
Further fine-grained tuning could improve robustness (albeit to a limited extent).

\subsection{Inner maximization}

\subsubsection{Number of optimization steps}

It is important that the attack used during training be strong enough.
Weak attacks tend to provide a false sense of security by allowing the trained network to use obfuscation as a defense mechanism \citep{qin_adversarial_2019}.
In this experiment, we study the effect of the attack strength on robustness.
Although recent work demonstrated that it is possible to train robust model with single-step attacks~\citep{wong_fast_2020}, it is generally accepted that the number of steps used for the inner optimization correlates with the strength of that optimization procedure (i.e., its ability to the find a minima close to the global minima).
Note, however, that more inner steps leads to increased training time.

\begin{table}[t]
\begin{center}
\caption{\footnotesize{Clean (without perturbations) and robust (under adversarial attack) accuracy obtained when increasing the number of steps used for the inner optimization, trained on \cifar (with and without additional unlabeled data) against \linf norm-bounded perturbations of size $\epsilon = 8/255$.}
\label{table:attack_steps}}
\resizebox{.95\textwidth}{!}{
\begin{tabular}{p{0.42\textwidth}|cc|cc}
    \hline
    \cellcolor{header} & \multicolumn{2}{c|}{\cellcolor{header} \cifar} & \multicolumn{2}{c}{\cellcolor{header} with \tinyimages} \Tstrut \\
    \cellcolor{header} \textsc{Setup} & \cellcolor{header} \textsc{Clean} & \cellcolor{header} \textsc{Robust} & \cellcolor{header} \textsc{Clean} & \cellcolor{header} \textsc{Robust} \Bstrut \\
    \hline
    \hline
    \multicolumn{5}{l}{\cellcolor{header} \textsc{Number of inner optimization steps}} \TBstrut \\
    \hline
    $K = 1, \alpha = 10/255$ & 70.31\% & 33.48\% & 76.64\% & 41.79\% \Tstrut \\
    $K = 2, \alpha = 5/255$ & 87.84\% & 47.75\% & 92.34\% & 53.93\% \\
    $K = 4, \alpha = 2.5/255$ & 87.98\% & 47.91\% & 91.80\% & 55.49\% \\
    $K = 8, \alpha = 0.007$ & 86.32\% & 50.05\% & 90.99\% & 57.93\% \\
    $K = 10, \alpha = 0.007$ & 84.85$\pm$1.20\% & 50.80$\pm$0.23\% & 90.93$\pm$0.25\% & 58.41$\pm$0.25\% \\
    $K = 16, \alpha = 0.007$ & 85.90\% & 51.31\% & 90.73\% & 58.87\% \Bstrut \\
    \hline
\end{tabular}
}
\end{center}
\end{table}

\paragraph{Results.}

In \autoref{table:attack_steps}, we vary the number of attack steps $K$ between $1$ and $16$.
We adapt the step-size by setting $\alpha$ to $\max(1.25 \epsilon / K, 0.007)$.
We observe that stronger attacks yield more robust models (with diminishing returns).
For example, increasing $K$ from $4$ to $8$, $8$ to $10$ and $10$ to $16$ improves robust accuracy by +2.59\%, +0.75\% and +0.51\% respectively (in the setting without additional data).

\paragraph{Key takeaways.}

Strong inner optimizations improve robustness (at the cost of increased training time).

\subsubsection{Inner maximization perturbation radius (continued)}

This section continues the evaluation made in \autoref{sec:inner_radius}.
In particular, we evaluate the effect of using a larger perturbation radius within TRADES~\citep{zhang_theoretically_2019} (\autoref{sec:inner_radius} explored this effect within classical adversarial training).

\begin{table}[t]
\begin{center}
\caption{\footnotesize{Clean (without perturbations) and robust (under adversarial attack) accuracy obtained on \cifar (with and without additional unlabeled data) against \linf norm-bounded perturbations of size $\epsilon = 8/255$ when using different perturbation radii for the inner maximization trained.}
\label{table:trades_epsilon}}
\resizebox{0.95\textwidth}{!}{
\begin{tabular}{p{0.6\textwidth}|cc|cc}
    \hline
    \cellcolor{header} & \multicolumn{2}{c|}{\cellcolor{header} \cifar} & \multicolumn{2}{c}{\cellcolor{header} with \tinyimages} \Tstrut \\
    \cellcolor{header} \textsc{Setup} & \cellcolor{header} \textsc{Clean} & \cellcolor{header} \textsc{Robust} & \cellcolor{header} \textsc{Clean} & \cellcolor{header} \textsc{Robust} \Bstrut \\
    \hline
    \hline
    \multicolumn{5}{l}{\cellcolor{header} \textsc{Perturbation radius used for training with TRADES}} \TBstrut \\
    \hline
    TRADES $\epsilon = 8/255$ \citep{zhang_theoretically_2019} & 82.74\% & 51.91\% & 88.36\% & 59.45\% \Tstrut \\
    TRADES $\epsilon = 8.8/255$ & 82.93\% & 52.80\% & 86.86\% & 59.21\% \\
    TRADES $\epsilon = 9.6/255$ & 81.85\% & 52.50\% & 85.79\% & 58.64\% \Bstrut \\
    \hline
\end{tabular}
}
\end{center}
\end{table}

\paragraph{Results.}
\autoref{table:trades_epsilon} shows the effect of increasing the perturbation radius $\epsilon$ by a factor $1.1\times$ and $1.2\times$ the original value of $8/255$ (during training, not during evaluation).
We notice that to the contrary of adversarial training, which benefits from increases perturbation radii, TRADES' performance is inconsistent (possibly because TRADES is already actively managing the trade-off with clean accuracy).

\paragraph{Key takeaways.}
When using TRADES, tuning the perturbation radius is not always beneficial.

\subsection{Additional unlabeled data}

\subsubsection{Ratio of labeled-to-unlabeled data per\\batch (continued)}

\begin{wrapfigure}{r}{0.5\textwidth}
\begin{center}
\vspace{-3.2cm}
\includegraphics[width=.45\textwidth]{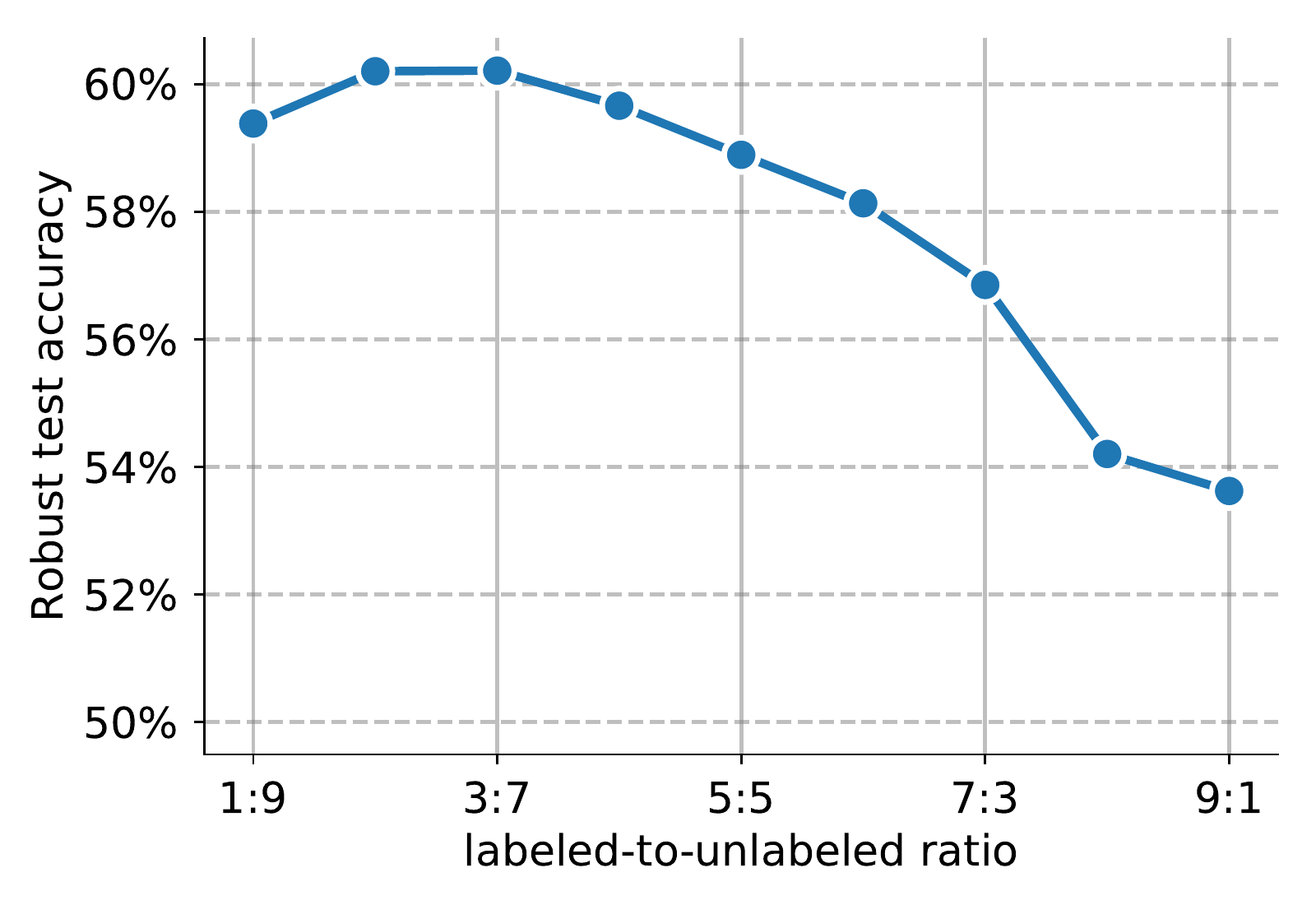}
\end{center}
\caption{Accuracy under \linf attacks of size $\epsilon = 8/255$ on \cifar as we vary the ratio of labeled-to-unlabeled data using our unlabeled datasets of 1M images.\label{fig:ratio_summary_unlabeled_app}}
\vspace{-1.5cm}
\end{wrapfigure}

This section continues the evaluation made in \autoref{sec:ratio_unlabeled}.
We evaluate the effect of varying the labeled-to-unlabeled data ratio on our largest unlabeled dataset consisting of 1M images.

\paragraph{Results.}

\autoref{fig:ratio_summary_unlabeled_app} shows the robust accuracy as we vary that ratio.
Similarly to \autoref{fig:ratio_summary_unlabeled_main} (which used the dataset from \citealp{carmon_unlabeled_2019}), we observe that giving slightly more importance to unlabeled data helps.
We find an identical optimal ratio of labeled-to-unlabeled data of 3:7, which provides a boost of +1.32\% over the 1:1 ratio.
When we use the dataset from \citet{carmon_unlabeled_2019}, the optimal ratio provides a boost of +0.95\% only.
This could indicate that larger gains in robustness are possible when using larger unlabeled datasets.

\paragraph{Key takeaways.}
Larger unlabeled datasets can provide larger improvements in robustness.

\subsubsection{Label smoothing}
\label{sec:ls}

As hinted in \autoref{sec:ratio_unlabeled}, we experiment  with  label  smoothing (for the additional unlabeled data).
Label smoothing should counteract the effect of the noisy labels resulting from the classifier used in the \emph{pseudo-labeling} process.
Label smoothing modifies one-hot labels $y$ by creating smoother targets $\hat{y} = (1 - \gamma) y + \gamma \vone$.
More specifically, we minimize the following loss
\begin{align}
\mathcal{L}_\vtheta^\textrm{AT-smooth} = \xent(f(\vx + \hat{\vdelta}; \vtheta), \hat{y}), ~~\mbox{ where }
\hat{\vdelta} \approx \argmax_{\vdelta \in \sS} \xent(f(\vx + \vdelta; \vtheta), y).
\end{align}

\begin{table}[t]
\begin{center}
\caption{\footnotesize{Clean (without perturbations) and robust (under adversarial attack) accuracy obtained for different label smoothing factors (on unlabeled data only) trained on \cifar (with and without additional unlabeled data) against \linf norm-bounded perturbations of size $\epsilon = 8/255$.}
\label{table:ls_unlabeled}}
\resizebox{.95\textwidth}{!}{
\begin{tabular}{p{0.42\textwidth}|cc|cc}
    \hline
    \cellcolor{header} & \multicolumn{2}{c|}{\cellcolor{header} \cifar} & \multicolumn{2}{c}{\cellcolor{header} with \tinyimages} \Tstrut \\
    \cellcolor{header} \textsc{Setup} & \cellcolor{header} \textsc{Clean} & \cellcolor{header} \textsc{Robust} & \cellcolor{header} \textsc{Clean} & \cellcolor{header} \textsc{Robust} \Bstrut \\
    \hline
    \hline
    \multicolumn{5}{l}{\cellcolor{header} \textsc{Label smoothing for the additional unlabeled data}} \TBstrut \\
    \hline
    $\gamma = 0$ & 84.85$\pm$1.20\% & 50.80$\pm$0.23\% & 90.93$\pm$0.25\% & 58.41$\pm$0.25\% \Tstrut \\
    $\gamma = 0.01$ & -- & -- & 91.02\% & 57.97\% \\
    $\gamma = 0.02$ & -- & -- & 91.35\% & 57.98\% \\
    $\gamma = 0.05$ & -- & -- & 91.08\% & 57.65\% \\
    $\gamma = 0.1$ & -- & -- & 91.20\% & 57.72\% \\
    $\gamma = 0.2$ & -- & -- & 90.83\% & 58.10\% \Bstrut \\
    \hline
\end{tabular}
}
\end{center}
\end{table}

\paragraph{Results.}
In \autoref{table:ls_unlabeled}, we apply label smoothing to the examples originating from the additional unlabeled dataset (from \citealp{carmon_unlabeled_2019}).
We vary $\gamma$ for these examples only (i.e., the labeled data from \cifar continues to use hard labels with $\gamma = 0$).
We observe that label smoothing is detrimental and that the resulting robust accuracy is inconsistent, without a clear correlation with $\gamma$.

\paragraph{Key takeaways.}

Applying label smoothing to the additional unlabeled data is detrimental to robustness.

\subsection{Other tricks}

\subsubsection{Batch size}

Using larger batch size not only influences resource utilization but also affects the optimization process.
Larger batches provide less noisy gradients (e.g., when using Stochastic Gradient Descent) and more precise batch statistics.
In classical adversarial training, it is common to let the batch statistics ``float'' as the inner optimization process is run.\footnote{\citet{xie_intriguing_2019} experimented with alternative batch normalization strategies.}
As such, batch size also influences the quality of the adversarial examples generated during training.
In this experiment, we vary the batch size and compensate for the loss of gradient noise by scaling the outer learning rate using the linear scaling
rule introduced by \citet{goyal2017accurate} (i.e., $\textrm{effective learning} = \max(\textrm{learning rate} \times \textrm{batch size} / 256, \textrm{learning rate})$).

\begin{table}[b]
\begin{center}
\caption{\footnotesize{Clean (without perturbations) and robust (under adversarial attack) accuracy obtained for batch sizes trained on \cifar (with and without additional unlabeled data) against \linf norm-bounded perturbations of size $\epsilon = 8/255$.}
\label{table:batchsize}}
\resizebox{.95\textwidth}{!}{
\begin{tabular}{p{0.42\textwidth}|cc|cc}
    \hline
    \cellcolor{header} & \multicolumn{2}{c|}{\cellcolor{header} \cifar} & \multicolumn{2}{c}{\cellcolor{header} with \tinyimages} \Tstrut \\
    \cellcolor{header} \textsc{Setup} & \cellcolor{header} \textsc{Clean} & \cellcolor{header} \textsc{Robust} & \cellcolor{header} \textsc{Clean} & \cellcolor{header} \textsc{Robust} \Bstrut \\
    \hline
    \hline
    \multicolumn{5}{l}{\cellcolor{header} \textsc{Batch size}} \TBstrut \\
    \hline
    128 & 84.85$\pm$1.20\% & 50.80$\pm$0.23\% & 90.13\% & 56.71\% \Tstrut \\
    256 & 83.39\% & 50.74\% & 90.71\% & 57.57\% \\
    512 & 83.31\% & 51.68\% & 90.96\% & 57.75\% \\
    1024 & 83.54\% & 50.60\% & 90.93$\pm$0.25\% & 58.41$\pm$0.25\% \Bstrut \\
    \hline
\end{tabular}
}
\end{center}
\end{table}

\paragraph{Results.}

\autoref{table:batchsize} shows the robust accuracy resulting from using different batch sizes.
We observe that our default batch size of $128$ is sub-optimal for the setting without additional unlabeled data: a batch size of $512$ improves robust accuracy by +0.88\%.
For the setting with additional unlabeled data, the largest batch size (i.e., $1024$) remains the best, as it improves on smaller batch sizes by at least +0.66\%.

\paragraph{Key takeaways.}

Training batch size has an effect on robustness.

\subsubsection{Data augmentation}

Data augmentation can reduce generalization error.
For image classification tasks, random flips, rotations and crops are commonly used~\citep{he2015deep}.
As is common for \cifar, in our baseline settings, we apply random translations by up to 4 pixels and random horizontal flips.
More sophisticated techniques such as \emph{Cutout}~\citep{devries2017improved} (which produces random occlusions) and \emph{mixup}~\citep{zhang2017mixup} (which linearly interpolates between two images) demonstrate compelling results on standard classification tasks.
However, both techniques are not very effective when used in conjunction with adversarial training~\citep{rice_overfitting_2020}.
In this experiment, we evaluate \emph{AutoAugment}~\citep{cubuk_autoaugment:_2018}, \emph{RandAugment}~\citep{cubuk2019randaugment} and the \emph{AugMix} augmentation~\citep{hendrycks2019augmix} (with their default settings).
We also evaluate the color augmentation scheme proposed by \citet{chen_simple_2020} (as part of the \emph{SimCLR} pipeline) by varying the \emph{color jittering} strength.
Irrespective of the strength, this color augmentation scheme always includes a random color drop (i.e., conversion to gray-scale) with a probability of 20\%.

\begin{table}[t]
\begin{center}
\caption{\footnotesize{Clean (without perturbations) and robust (under adversarial attack) accuracy obtained for different data augmentation schemes trained on \cifar (with and without additional unlabeled data) against \linf norm-bounded perturbations of size $\epsilon = 8/255$.}
\label{table:augment}}
\resizebox{.95\textwidth}{!}{
\begin{tabular}{l|cc|cc}
    \hline
    \cellcolor{header} & \multicolumn{2}{c|}{\cellcolor{header} \cifar} & \multicolumn{2}{c}{\cellcolor{header} with \tinyimages} \Tstrut \\
    \cellcolor{header} \textsc{Setup} & \cellcolor{header} \textsc{Clean} & \cellcolor{header} \textsc{Robust} & \cellcolor{header} \textsc{Clean} & \cellcolor{header} \textsc{Robust} \Bstrut \\
    \hline
    \hline
    \multicolumn{5}{l}{\cellcolor{header} \textsc{Data augmentation}} \TBstrut \\
    \hline
    default = translate(4) + flip(0.5) & 84.85$\pm$1.20\% & 50.80$\pm$0.23\% & 90.93$\pm$0.25\% & 58.41$\pm$0.25\% \Tstrut \\
    \emph{AutoAugment}~\citep{cubuk_autoaugment:_2018} & 86.24\% & 50.37\% & 90.21\% & 56.57\% \\
    \emph{RandAugment}~\citep{cubuk2019randaugment} & 85.01\% & 47.37\% & 88.93\% & 52.48\% \\
    \emph{AugMix}~\citep{hendrycks2019augmix} & 82.05\% & 50.12\% & 89.67\% & 56.62\% \\
    default + color-jitter(0)~\citep{chen_simple_2020} & 84.29\% & 51.49\% & 90.53\% & 58.70\% \\
    default + color-jitter(0.1)~\citep{chen_simple_2020} & 84.19\% & 50.87\% & 91.05\% & 58.27\% \\
    default + color-jitter(0.2)~\citep{chen_simple_2020} & 83.80\% & 51.10\% & 90.97\% & 58.43\% \\
    default + color-jitter(0.3)~\citep{chen_simple_2020} & 82.87\% & 50.89\% & 91.05\% & 57.98\% \\
    default + color-jitter(0.4)~\citep{chen_simple_2020} & 83.42\% & 50.01\% & 91.24\% & 57.04\% \\
    default + color-jitter(0.5)~\citep{chen_simple_2020} & 83.43\% & 49.51\% & 90.79\% & 57.20\% \Bstrut \\
    \hline
\end{tabular}
}
\end{center}
\end{table}

\paragraph{Results.}
\autoref{table:augment} summarizes the results.
We observe that \emph{AutoAugment}, \emph{RandAugment} and \emph{AugMix}, which have mainly been tuned for \imagenet, reduce robust accuracy.
These techniques would require further fine-tuning to be competitive with the simplest augmentation scheme (i.e., random translation by 4 pixels).
Furthermore, we observe that increasing the strength of \emph{color jittering} correlates negatively with robustness, as robust accuracy drops by -1.98\% and -1.50\% in the settings without and with additional data, respectively.
Finally, we note that randomly dropping color does improve robustness in both settings (i.e., +0.69\% and +0.29\% with respect to the baselines).

\paragraph{Key takeaways.}
Data augmentation schemes that perform well for standard classification tasks do not necessarily improve robust generalization.

\subsubsection{Label smoothing}

In this section, we complement the label smoothing experiment done in \autoref{sec:ls}.
To the contrary of \autoref{sec:ls}, which explored label smoothing for unlabeled data only, this section explores label smoothing for the labeled data.

\begin{table}[b]
\begin{center}
\caption{\footnotesize{Clean (without perturbations) and robust (under adversarial attack) accuracy obtained for different label smoothing factors (on labeled data only) trained on \cifar (with and without additional unlabeled data) against \linf norm-bounded perturbations of size $\epsilon = 8/255$.}
\label{table:ls_labeled}}
\resizebox{.95\textwidth}{!}{
\begin{tabular}{p{0.42\textwidth}|cc|cc}
    \hline
    \cellcolor{header} & \multicolumn{2}{c|}{\cellcolor{header} \cifar} & \multicolumn{2}{c}{\cellcolor{header} with \tinyimages} \Tstrut \\
    \cellcolor{header} \textsc{Setup} & \cellcolor{header} \textsc{Clean} & \cellcolor{header} \textsc{Robust} & \cellcolor{header} \textsc{Clean} & \cellcolor{header} \textsc{Robust} \Bstrut \\
    \hline
    \hline
    \multicolumn{5}{l}{\cellcolor{header} \textsc{Label smoothing for the labeled data}} \TBstrut \\
    \hline
    $\gamma = 0$ & 84.85$\pm$1.20\% & 50.80$\pm$0.23\% & 90.93$\pm$0.25\% & 58.41$\pm$0.25\% \Tstrut \\
    $\gamma = 0.01$ & 84.39\% & 50.65\% & 90.50\% & 58.86\% \\ 
    $\gamma = 0.02$ & 82.88\% & 51.13\% & 91.11\% & 58.46\% \\ 
    $\gamma = 0.05$ & 83.85\% & 50.91\% & 90.82\% & 58.34\% \\ 
    $\gamma = 0.1$  & 84.09\% & 50.26\% & 91.28\% & 58.66\% \\
    $\gamma = 0.2$  & 83.49\% & 51.25\% & 90.70\% & 58.89\% \Bstrut \\
    \hline
\end{tabular}
}
\end{center}
\end{table}

\paragraph{Results.}
\autoref{table:ls_labeled} shows the results.
We do not observe any clear correlation between label smoothing and robust accuracy.
In particular, setting the label smoothing factor $\gamma$ to 0.02 and 0.2 seems helpful (in both data settings), while setting it to 0.05 or 0.1 seems detrimental in at least one of the two data settings.

\paragraph{Key takeaways.}
Applying label smoothing to the labeled data has minor effects on robustness.

\section{Comparison with concurrent works}

Independently of our work, \citet{pang2020bag}\footnote{Accepted to the 2021 Conference on Learning Representations and available on ArXiv on October 1st, 2020.} also investigate the limits of current approaches to adversarial training.
We also highlight concurrent works by \citet{chen2021robust}\footnote{Accepted to the 2021 Conference on Learning Representations and available on OpenReview on September 28th, 2020.} and \citet{wu2020adversarial}\citet{wu2020adversarial}\footnote{Accepted at the 2020 Conference on Neural Information Processing Systems, but only available in its final form (with their latest results) from ArXiv on October 13th, 2020.}.
We briefly summarize both these works here for clarity and completeness.

\paragraph{\citet{pang2020bag}.}

This work is the closest in essence to ours.
\citet{pang2020bag} make a thorough literature review and observe that different papers on adversarial robustness differ in their hyper-parameters (despite using similar techniques).
While they analyze some properties that we also analyze in this manuscript (such as training batch size, label smoothing, weight decay, activation functions), they also complement our analyses with experiments on early stopping and perturbation radius warm-up, optimizers, model architectures beyond Wide-ResNets and batch normalization.
The combination of their findings applied to a \wrn-34-20 reaches 54.39\% robust accuracy without additional unlabeled data (in comparison, our \wrn-34-20 reaches a robust accuracy of 56.86\%).
Their study hints at further improvements by more finely tuning the weight decay.

\paragraph{\citet{wu2020adversarial}.}

In their manuscript, \citet{wu2020adversarial} explore \emph{adversarial weight perturbation} as way to improve robust generalization.
Their technique interleaves model weight perturbations with example perturbations (within a single training step).
They observe that the resulting loss landscapes become flatter and, as a result, robustness improves.
Using this approach, \citet{wu2020adversarial} improve robust accuracy by significant margins reaching 56.17\% without additional data.
By combining our findings with their technique, we expect that further improvements are possible (although we posit that model weight averaging may already have a similar effect).

\paragraph{\citet{chen2021robust}.}

Simultaneously to us, \citet{chen2021robust} discovered that model weight averaging can significantly improve robustness on a wide range of models and datasets.
They argue (similarly to \citealp{wu2020adversarial}) that WA leads to a flatter adversarial loss landscape, and thus a smaller robust generalization gap.
In addition to matching our experimental results on WA, they provide a deeper, noteworthy analysis.

\clearpage
\section{Additional detailed results}
\label{sec:additional_tables}

\begin{table}[h]
\begin{center}
\caption{\footnotesize{Clean (without perturbations) and robust (under adversarial attack) accuracy obtained by different network sizes trained on \cifar (with and without additional unlabeled data) against \linf norm-bounded perturbations of size $\epsilon = 8/255$.}
\label{table:width_depth}}
\resizebox{.95\textwidth}{!}{
\begin{tabular}{p{0.42\textwidth}|cc|cc}
    \hline
    \cellcolor{header} & \multicolumn{2}{c|}{\cellcolor{header} \cifar} & \multicolumn{2}{c}{\cellcolor{header} with \tinyimages} \Tstrut \\
    \cellcolor{header} \textsc{Setup} & \cellcolor{header} \textsc{Clean} & \cellcolor{header} \textsc{Robust} & \cellcolor{header} \textsc{Clean} & \cellcolor{header} \textsc{Robust} \Bstrut \\
    \hline
    \hline
    \multicolumn{5}{l}{\cellcolor{header} \textsc{Network Width and Depth}} \TBstrut \\
    \hline
    \wrn-28-10 & 84.85$\pm$1.20\% & 50.80$\pm$0.23\% & 90.93$\pm$0.25\% & 58.41$\pm$0.25\% \Tstrut \\
    \wrn-28-15 & 86.91\% & 51.79\% & 91.71\% & 59.81\% \\
    \wrn-28-20 & 86.87\% & 51.78\% & 91.63\% & 59.90\% \\
    \wrn-34-10 & 86.42\% & 51.18\% & 91.45\% & 58.52\% \\
    \wrn-34-15 & 86.40\% & 51.12\% & 91.84\% & 59.61\% \\
    \wrn-34-20 & 87.32\% & 52.91\% & 92.14\% & 60.90\% \\
    \wrn-40-10 & 86.27\% & 51.89\% & 91.47\% & 59.47\% \\
    \wrn-40-15 & 86.94\% & 52.18\% & 91.66\% & 60.44\% \\
    \wrn-40-20 & 86.62\% & 53.19\% & 91.98\% & 60.97\% \\
    \wrn-46-10 & 85.98\% & 52.02\% & 91.71\% & 59.40\% \\
    \wrn-46-15 & 86.73\% & 52.74\% & 91.78\% & 60.56\% \\
    \wrn-46-20 & 87.22\% & 53.24\% & 92.08\% & 61.14\% \Bstrut \\
    \hline
\end{tabular}
}
\end{center}
\end{table}

\begin{table}[h]
\begin{center}
\caption{\footnotesize{Clean (without perturbations) and robust (under adversarial attack) accuracy obtained by weight averaging decay values trained on \cifar (with and without additional unlabeled data) against \linf norm-bounded perturbations of size $\epsilon = 8/255$.}
\label{table:ema}}
\resizebox{.95\textwidth}{!}{
\begin{tabular}{p{0.42\textwidth}|cc|cc}
    \hline
    \cellcolor{header} & \multicolumn{2}{c|}{\cellcolor{header} \cifar} & \multicolumn{2}{c}{\cellcolor{header} with \tinyimages} \Tstrut \\
    \cellcolor{header} \textsc{Setup} & \cellcolor{header} \textsc{Clean} & \cellcolor{header} \textsc{Robust} & \cellcolor{header} \textsc{Clean} & \cellcolor{header} \textsc{Robust} \Bstrut \\
    \hline
    \hline
    \multicolumn{5}{l}{\cellcolor{header} \textsc{Model Weight Averaging}} \TBstrut \\
    \hline
    No weight averaging & 84.85$\pm$1.20\% & 50.80$\pm$0.23\% & 90.93$\pm$0.25\% & 58.41$\pm$0.25\% \Tstrut \\
    Decay parameter $\tau=0.95$ & 85.48\% & 50.46\% & 91.07\% & 58.12\% \\
    $\tau=0.9625$ & 85.80\% & 50.53\% & 91.28\% & 58.23\% \\
    $\tau=0.975$ & 86.34\% & 50.65\% & 91.03\% & 58.37\% \\
    $\tau=0.9875$ & 83.47\% & 51.45\% & 91.09\% & 58.33\% \\
    $\tau=0.99$ & 83.41\% & 51.37\% & 91.51\% & 58.69\% \\
    $\tau=0.9925$ & 86.08\% & 51.44\% & 91.23\% & 58.74\% \\
    $\tau=0.995$ & 84.41\% & 52.10\% & 90.28\% & 59.14\% \\
    $\tau=0.9975$ & 84.91\% & 52.31\% & 91.28\% & 58.19\% \\
    $\tau=0.999$ & 84.62\% & 52.21\% & 91.66\% & 57.16\% \Bstrut \\
    \hline
\end{tabular}
}
\end{center}
\end{table}

\begin{table}[h]
\begin{center}
\caption{\footnotesize{Clean (without perturbations) and robust (under adversarial attack) accuracy obtained by different activations trained on \cifar (with and without additional unlabeled data) against \linf norm-bounded perturbations of size $\epsilon = 8/255$.}
\label{table:activation}}
\resizebox{.95\textwidth}{!}{
\begin{tabular}{p{0.42\textwidth}|cc|cc}
    \hline
    \cellcolor{header} & \multicolumn{2}{c|}{\cellcolor{header} \cifar} & \multicolumn{2}{c}{\cellcolor{header} with \tinyimages} \Tstrut \\
    \cellcolor{header} \textsc{Setup} & \cellcolor{header} \textsc{Clean} & \cellcolor{header} \textsc{Robust} & \cellcolor{header} \textsc{Clean} & \cellcolor{header} \textsc{Robust} \Bstrut \\
    \hline
    \hline
    \multicolumn{5}{l}{\cellcolor{header} \textsc{Activation}} \TBstrut \\
    \hline
    ReLU~\citep{nair2010rectified} & 84.85$\pm$1.20\% & 50.80$\pm$0.23\% & 90.93$\pm$0.25\% & 58.41$\pm$0.25\% \Tstrut \\
    Swish/SiLU~\citep{hendrycks2016gaussian} & 85.60\% & 51.40\% & 91.03\% & 59.54\% \\
    Leaky ReLU~\citep{maas2013rectifier} & 85.21\% & 51.00\% & 90.77\% & 58.55\% \\
    ELU~\citep{clevert2015fast} & 81.87\% & 49.20\% & 89.13\% & 56.48\% \\
    Softplus & 83.59\% & 50.61\% & 89.87\% & 57.83\% \\
    GELU~\citep{hendrycks2016gaussian} & 85.67\% & 50.66\% & 91.09\% & 59.37\% \Bstrut \\
    \hline
\end{tabular}
}
\end{center}
\end{table}

\begin{table}[h]
\begin{center}
\caption{\footnotesize{Clean (without perturbations) and robust (under adversarial attack) accuracy obtained using ratios of labeled-to-unlabeled data on \cifar and \tinyimages against \linf norm-bounded perturbations of size $\epsilon = 8/255$.}
\label{table:ratio}}
\resizebox{.95\textwidth}{!}{
\begin{tabular}{p{0.42\textwidth}|cc|cc}
    \hline
    \cellcolor{header} & \multicolumn{2}{c|}{\cellcolor{header} \cifar} & \multicolumn{2}{c}{\cellcolor{header} with \tinyimages} \Tstrut \\
    \cellcolor{header} \textsc{Setup} & \cellcolor{header} \textsc{Clean} & \cellcolor{header} \textsc{Robust} & \cellcolor{header} \textsc{Clean} & \cellcolor{header} \textsc{Robust} \Bstrut \\
    \hline
    \hline
    \multicolumn{5}{l}{\cellcolor{header} \textsc{Weighting of Unlabeled Data} (ratio of labeled-to-unlabeled data per batch)} \TBstrut \\
    \hline
    10:0 (more labeled data) & 84.85$\pm$1.20\% & 50.80$\pm$0.23\% & -- & -- \Tstrut \\
    9:1  & -- & -- & 88.60\% & 51.79\% \\
    8:2 & -- & -- & 86.74\% & 53.27\% \\
    7:3 & -- & -- & 90.37\% & 56.58\% \\
    6:4 & -- & -- & 90.85\% & 57.19\% \\
    5:5 & -- & -- & 90.93$\pm$0.25\% & 58.41$\pm$0.25\% \\
    4:6 & -- & -- & 91.06\% & 59.04\% \\
    3:7 & -- & -- & 90.99\% & 59.36\% \\
    2:8 & -- & -- & 90.69\% & 59.13\% \\
    1:9 (more unlabeled data) & -- & -- & 90.57\% & 58.06\% \Bstrut \\
    \hline
\end{tabular}
}
\end{center}
\end{table}

\begin{table}[t]
\begin{center}
\caption{\footnotesize{Clean (without perturbations) and robust (under adversarial attack) accuracy obtained for different weight decays trained on \cifar (with and without additional unlabeled data) against \linf norm-bounded perturbations of size $\epsilon = 8/255$.}
\label{table:l2_reg}}
\resizebox{.95\textwidth}{!}{
\begin{tabular}{p{0.42\textwidth}|cc|cc}
    \hline
    \cellcolor{header} & \multicolumn{2}{c|}{\cellcolor{header} \cifar} & \multicolumn{2}{c}{\cellcolor{header} with \tinyimages} \Tstrut \\
    \cellcolor{header} \textsc{Setup} & \cellcolor{header} \textsc{Clean} & \cellcolor{header} \textsc{Robust} & \cellcolor{header} \textsc{Clean} & \cellcolor{header} \textsc{Robust} \Bstrut \\
    \hline
    \hline
    \multicolumn{5}{l}{\cellcolor{header} \textsc{Number of training epochs}} \TBstrut \\
    \hline
    $0$ & 78.69\% & 45.74\% & 88.82\% & 53.58\% \Tstrut \\
    $5\cdot10^{-5}$ & 84.95\% & 49.34\% & 90.98\% & 56.31\% \\
    $1\cdot10^{-4}$ & 85.04\% & 50.72\% & 89.99\% & 57.40\% \\
    $5\cdot10^{-4}$ & 84.85$\pm$1.20\% & 50.80$\pm$0.23\% & 90.93$\pm$0.25\% & 58.41$\pm$0.25\% \\
    $1\cdot10^{-3}$ & 85.16\% & 50.44\% & 90.61\% & 58.41\% \\
    $5\cdot10^{-3}$ & 79.47\% & 47.61\% & 86.60\% & 55.10\% \Bstrut \\
    \hline
\end{tabular}
}
\end{center}
\end{table}

\clearpage
\section{Loss landscape analysis}

In this section, we analyze the adversarial loss landscapes of two of our best models trained without and with additional data against \linf norm-bounded perturbations of size $8/255$ on \cifar.
As a comparison, we also show the loss landscapes of the model trained by \citet{carmon_unlabeled_2019}.
This analysis complements the black-box Square attack~\cite{andriushchenko_square_2019} used by \autoattack.
To generate a loss landscape, we vary the network input along the linear space defined by the worse perturbation found by \pgd{40} ($u$ direction) and a random Rademacher direction ($v$ direction).
The $u$ and $v$ axes represent the magnitude of the perturbation added in each of these directions respectively and the $z$ axis is the adversarial margin loss~\citep{carlini_towards_2017}: $z_y - \max_{i \neq y} z_i$ (i.e., a misclassification occurs when this value falls below zero).

\begin{figure*}[h]
\centering
\begin{subfigure}{0.32\textwidth}
\centering
\includegraphics[width=\linewidth]{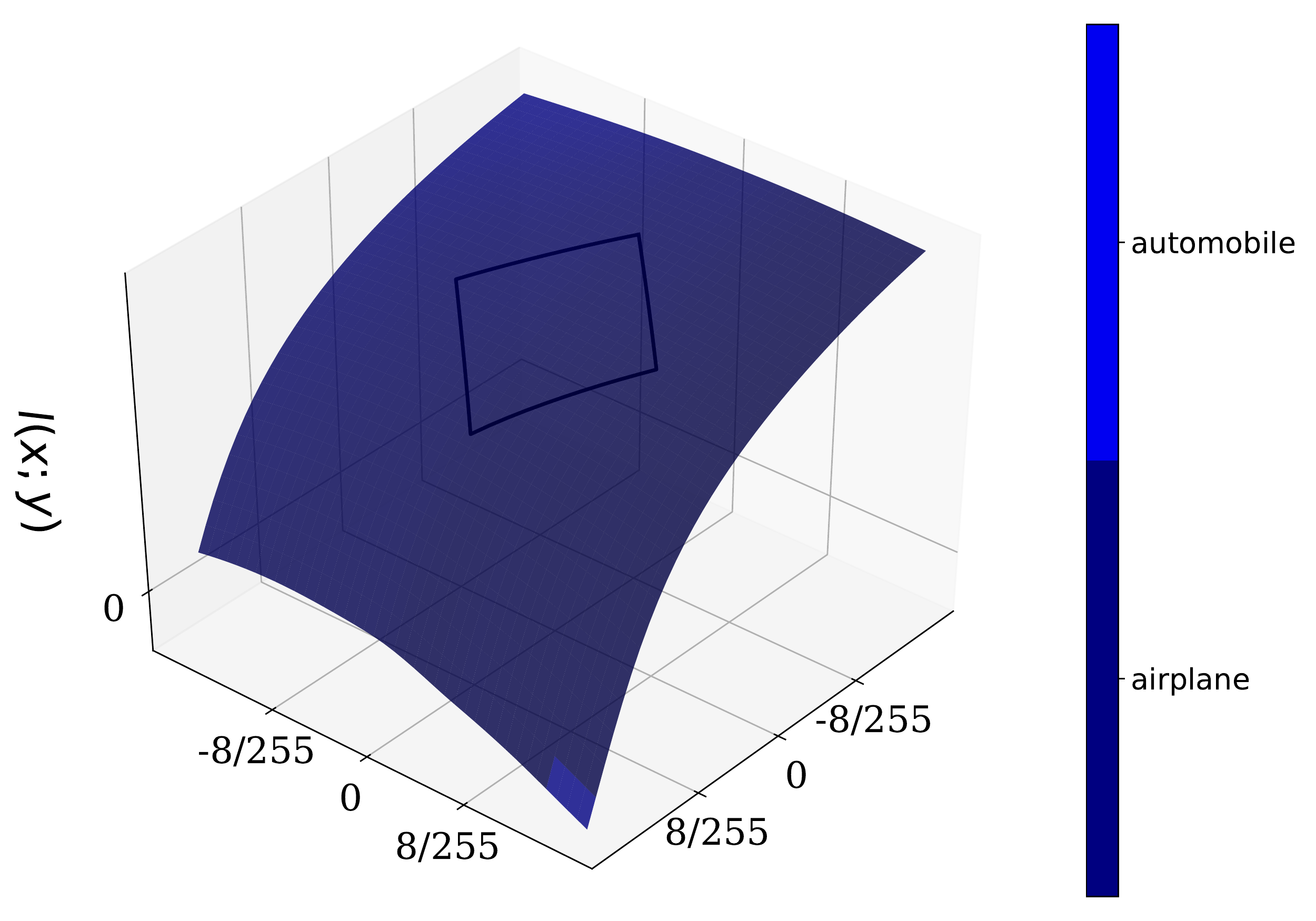}
\caption{airplane}
\end{subfigure}
\begin{subfigure}{0.32\textwidth}
\centering
\includegraphics[width=\linewidth]{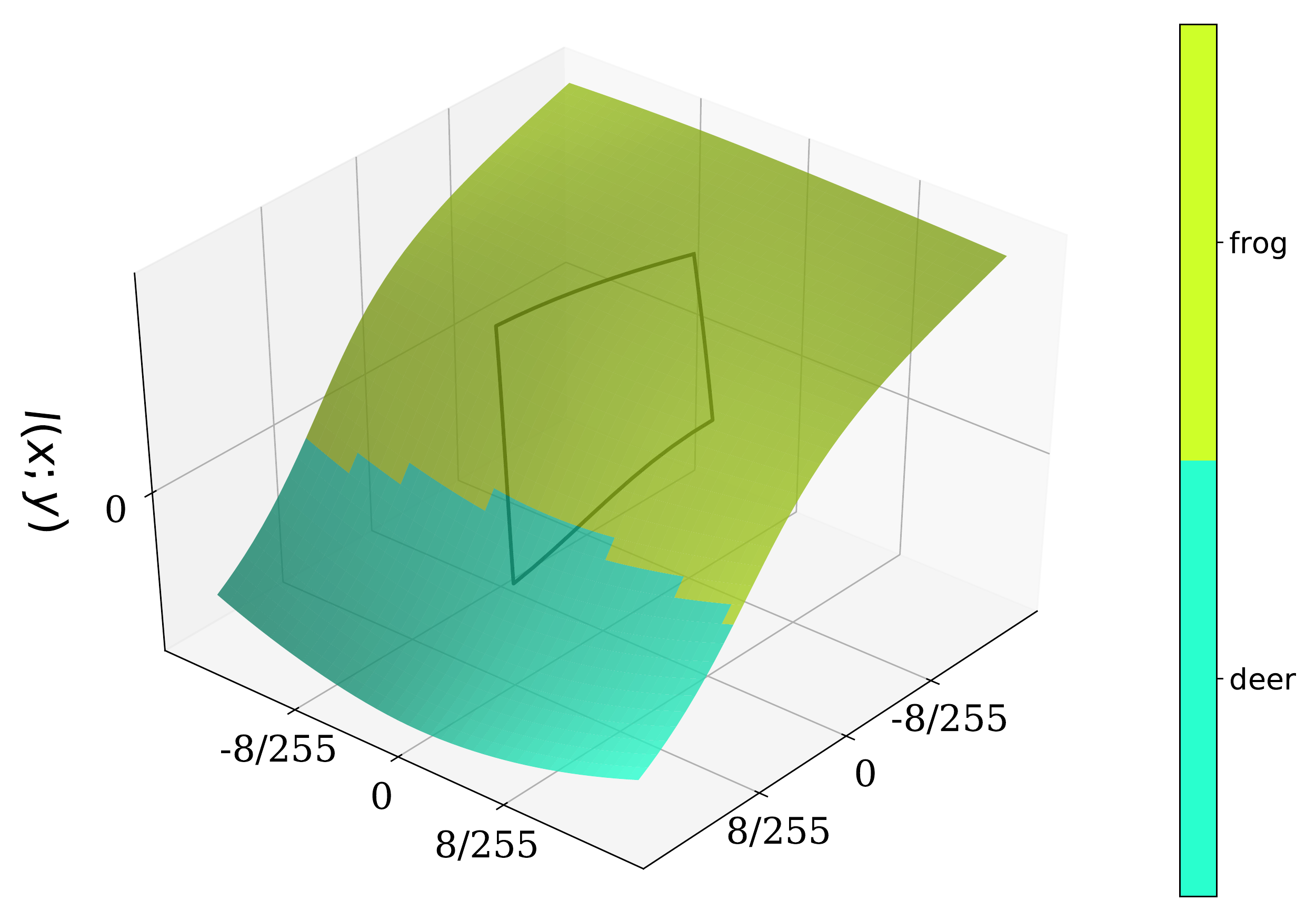}
\caption{frog}
\end{subfigure}
\begin{subfigure}{0.32\textwidth}
\centering
\includegraphics[width=\linewidth]{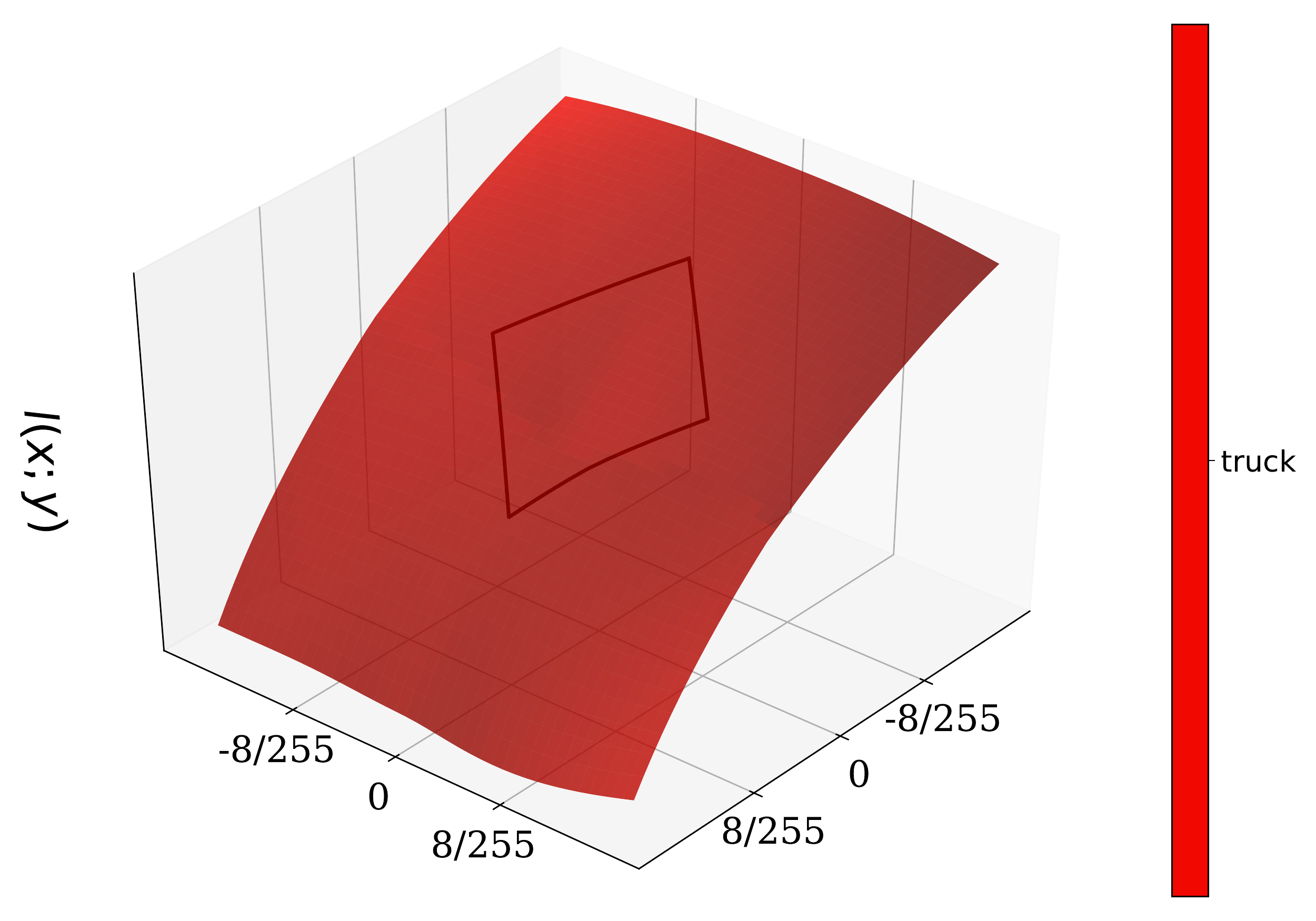}
\caption{truck}
\end{subfigure}
\begin{subfigure}{0.32\textwidth}
\centering
\includegraphics[width=\linewidth]{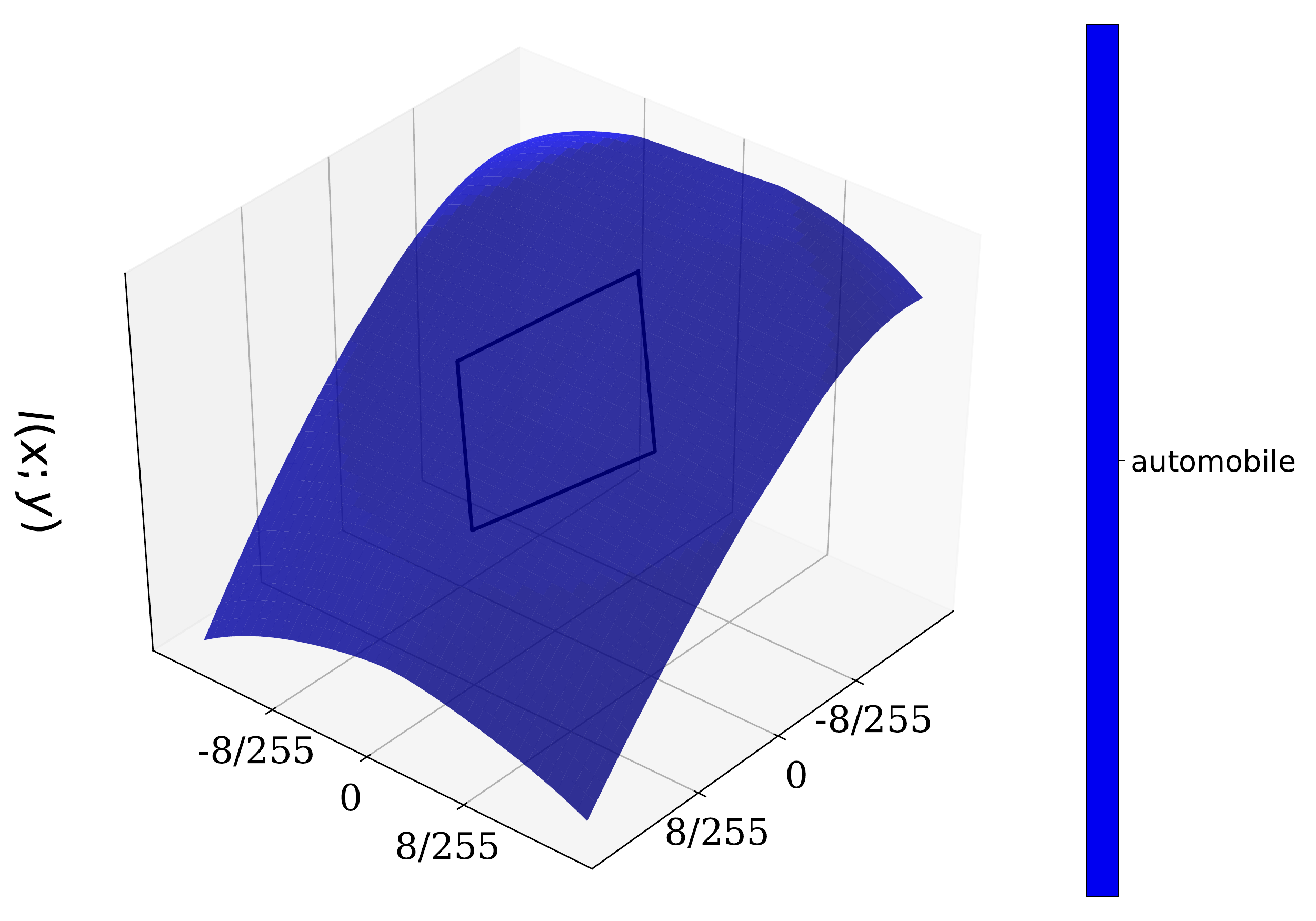}
\caption{automobile}
\end{subfigure}
\begin{subfigure}{0.32\textwidth}
\centering
\includegraphics[width=\linewidth]{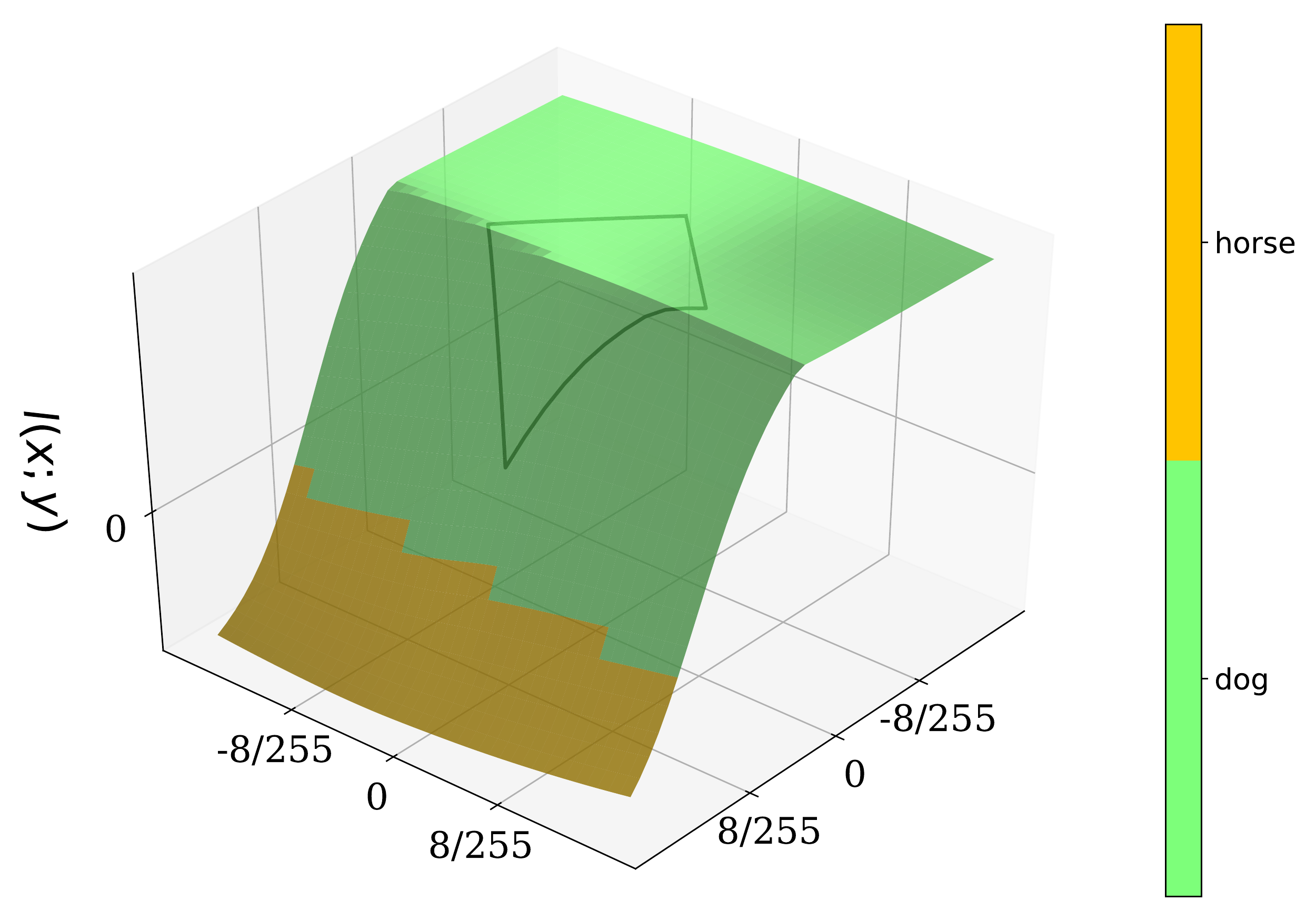}
\caption{dog}
\end{subfigure}
\caption{Loss landscapes around different \cifar test images.
It is generated by varying the input to the model, starting from the original input image toward either the worst attack found using \pgd{40} ($u$ direction) or a random Rademacher direction ($v$ direction). The loss used for these plots is the margin loss $z_y - \max_{i \neq y} z_i$ (i.e., a misclassification occurs when this value falls below zero). The model used is our \wrn-70-16 trained with additional unlabeled data against \linf norm-bounded perturbations of size $8/255$ on \cifar. The diamond-shape represents the projected \linf ball of size $\epsilon = 8/255$ around the nominal image.}
\label{fig:linf_data_landscapes}
\end{figure*}

\begin{figure*}[h]
\centering
\begin{subfigure}{0.32\textwidth}
\centering
\includegraphics[width=\linewidth]{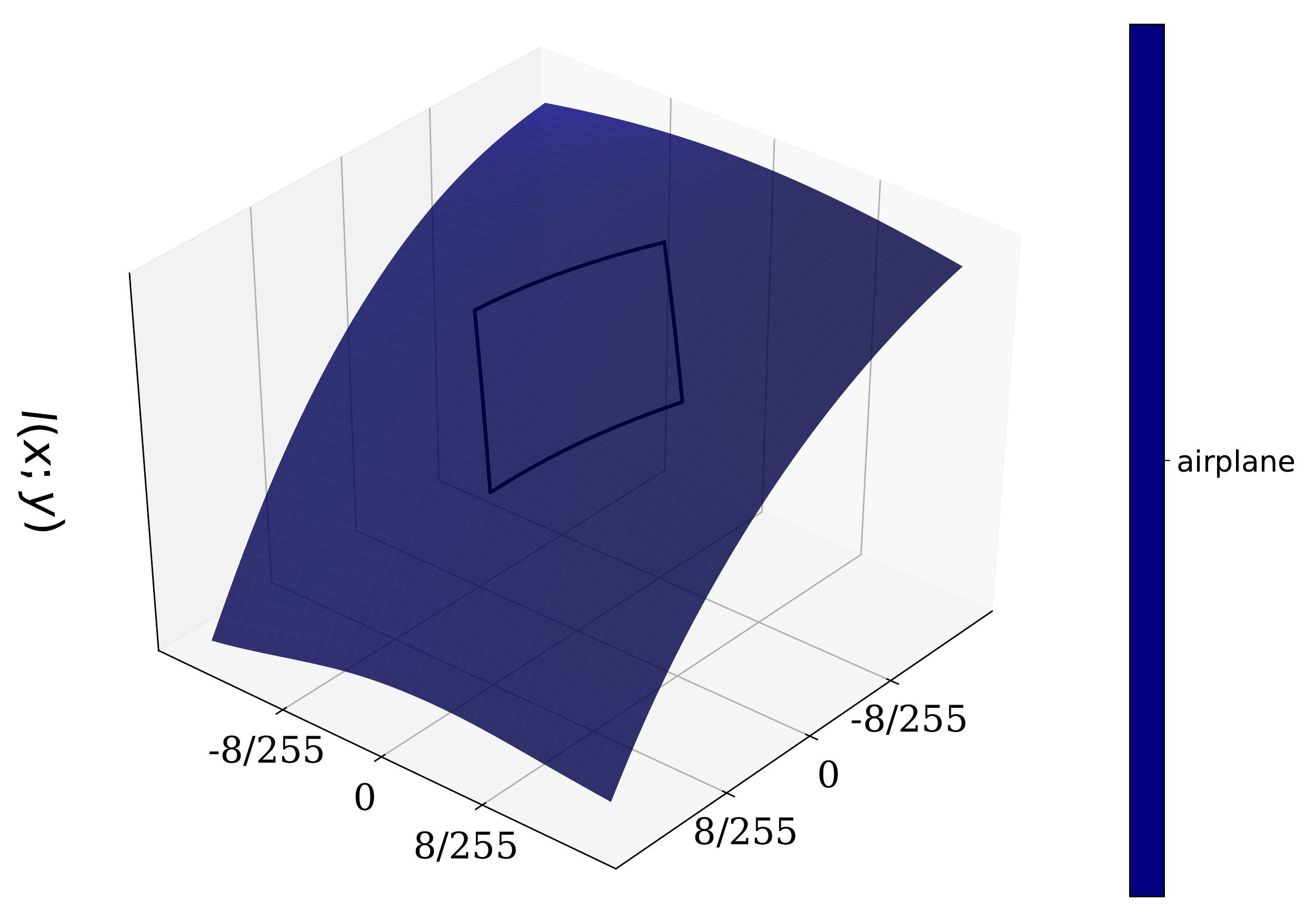}
\caption{airplane}
\end{subfigure}
\begin{subfigure}{0.32\textwidth}
\centering
\includegraphics[width=\linewidth]{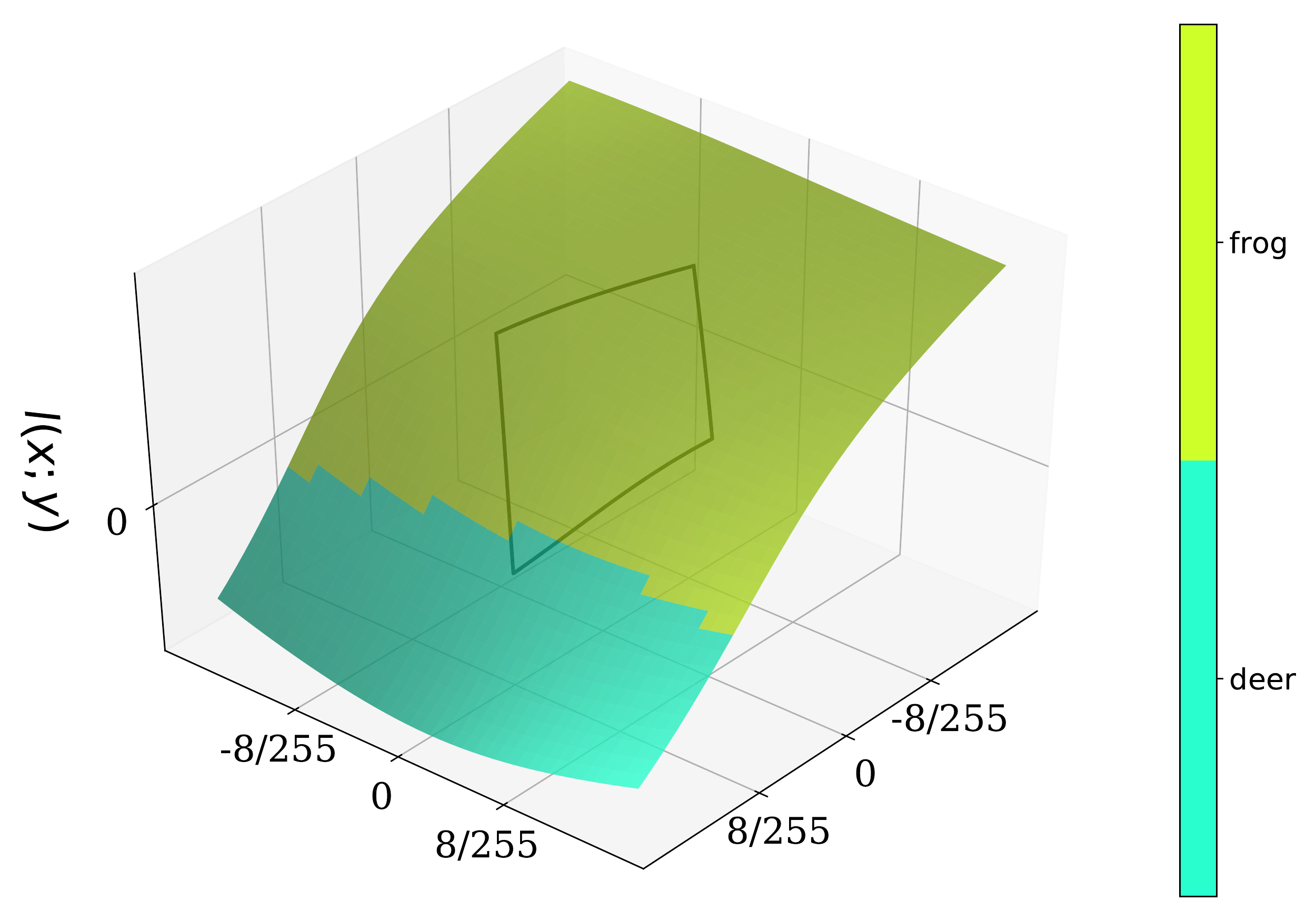}
\caption{frog}
\end{subfigure}
\begin{subfigure}{0.32\textwidth}
\centering
\includegraphics[width=\linewidth]{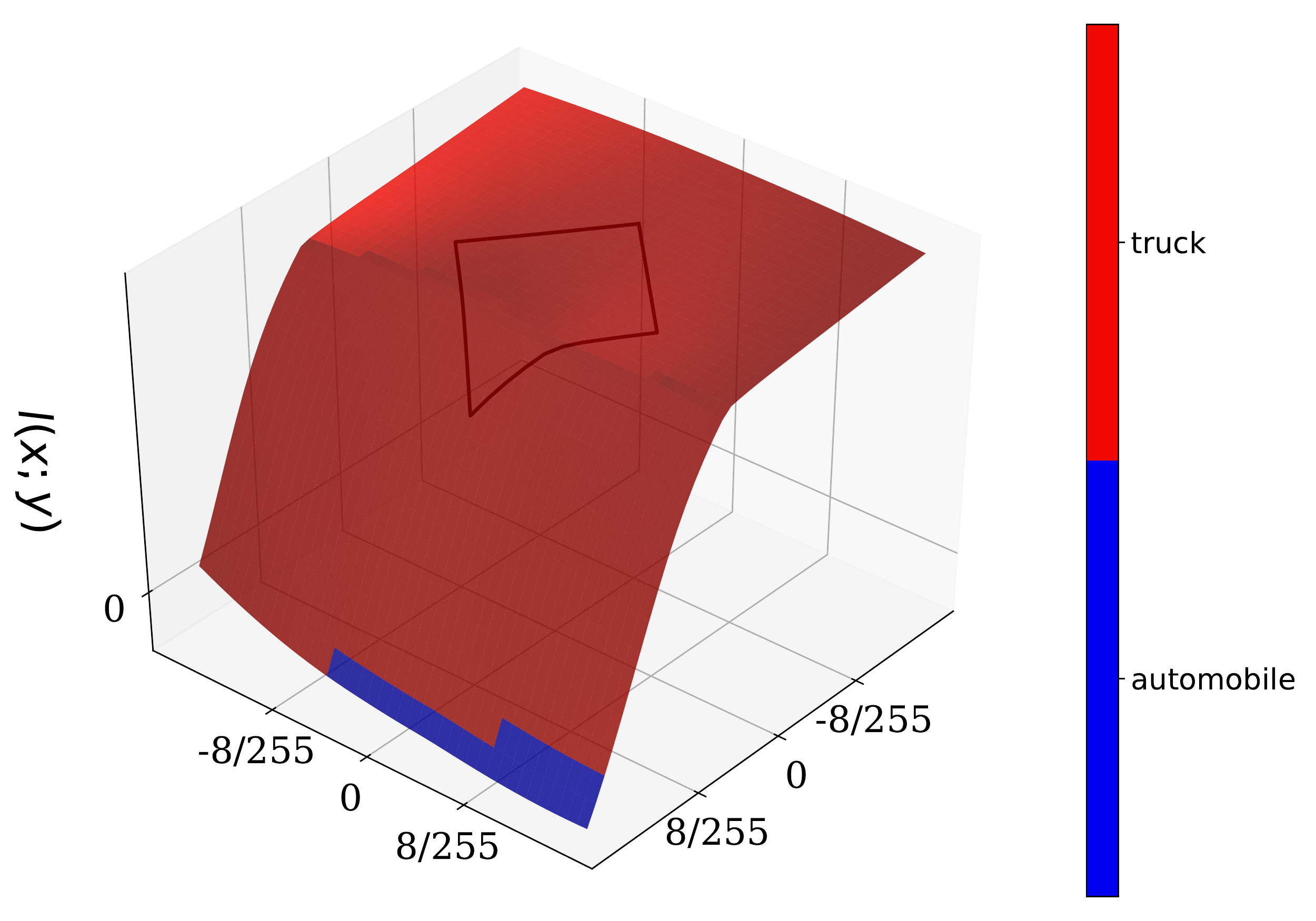}
\caption{truck}
\end{subfigure}
\begin{subfigure}{0.32\textwidth}
\centering
\includegraphics[width=\linewidth]{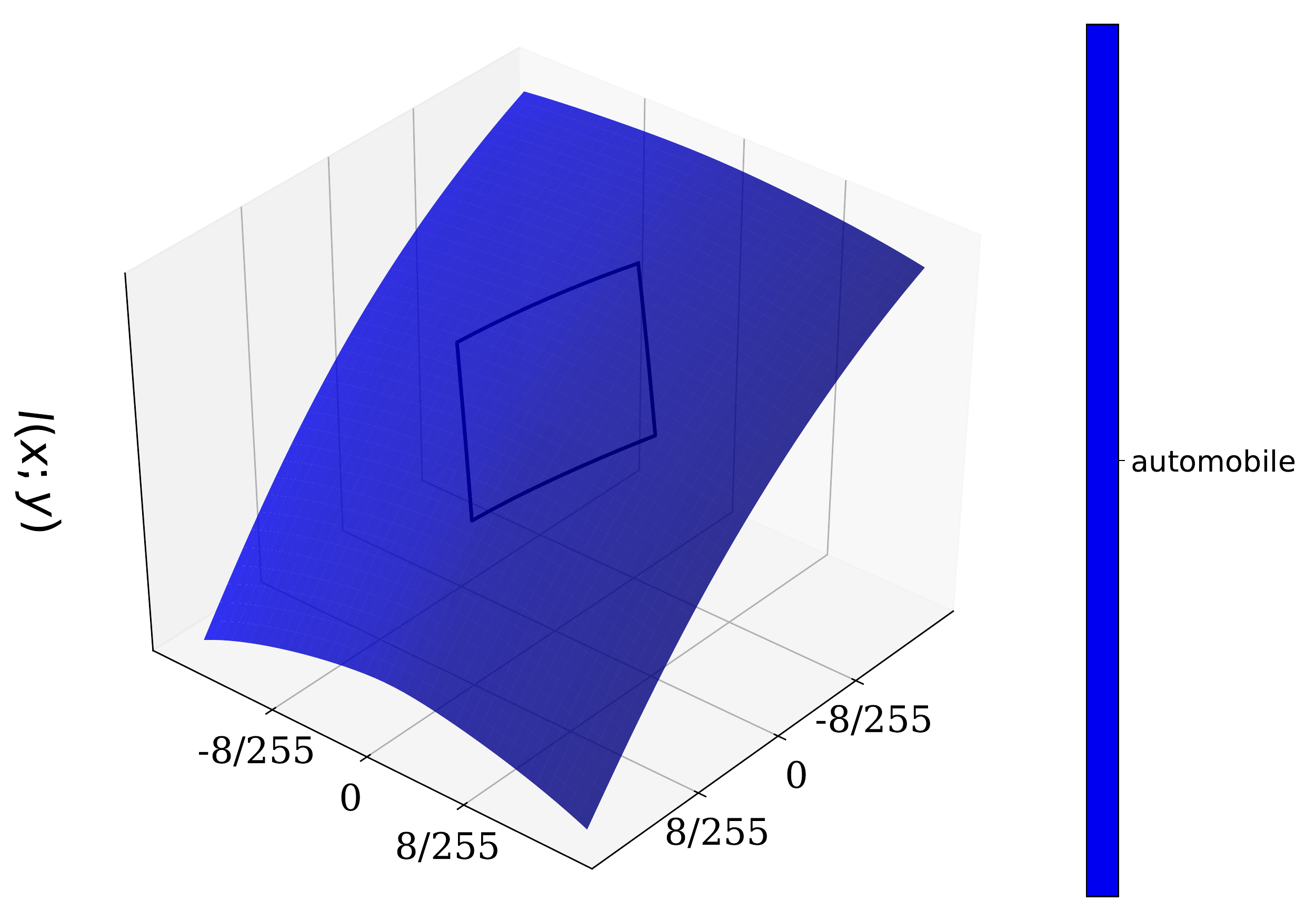}
\caption{automobile}
\end{subfigure}
\begin{subfigure}{0.32\textwidth}
\centering
\includegraphics[width=\linewidth]{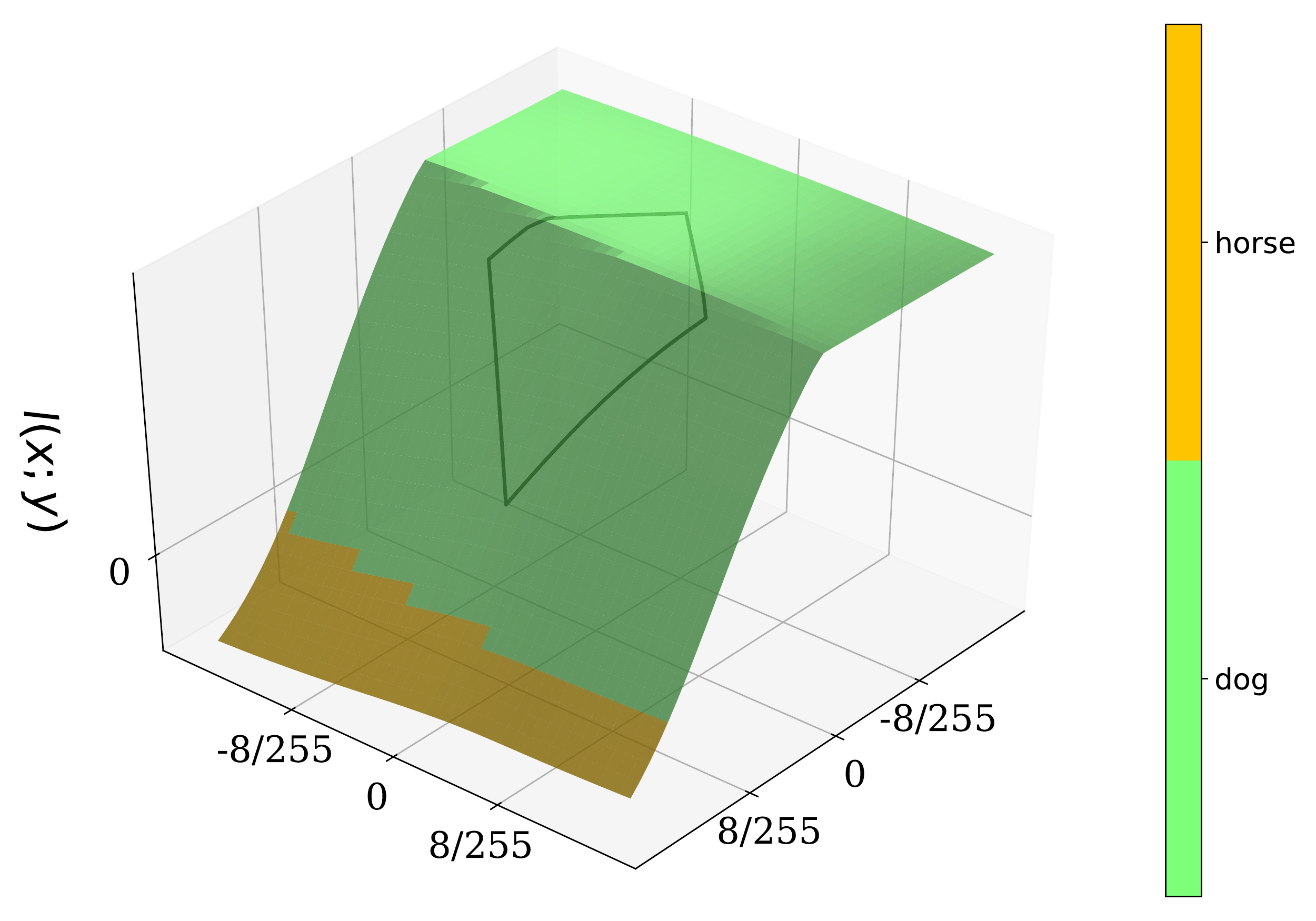}
\caption{dog}
\end{subfigure}
\caption{Loss landscapes around different \cifar test images.
It is generated by varying the input to the model, starting from the original input image toward either the worst attack found using \pgd{40} ($u$ direction) or a random Rademacher direction ($v$ direction). The loss used for these plots is the margin loss $z_y - \max_{i \neq y} z_i$ (i.e., a misclassification occurs when this value falls below zero). The model used is our \wrn-70-16 trained without additional unlabeled data against \linf norm-bounded perturbations of size $8/255$ on \cifar. The diamond-shape represents the projected \linf ball of size $\epsilon = 8/255$ around the nominal image.}
\label{fig:linf_landscapes}
\end{figure*}

\begin{figure*}[h]
\centering
\begin{subfigure}{0.32\textwidth}
\centering
\includegraphics[width=\linewidth]{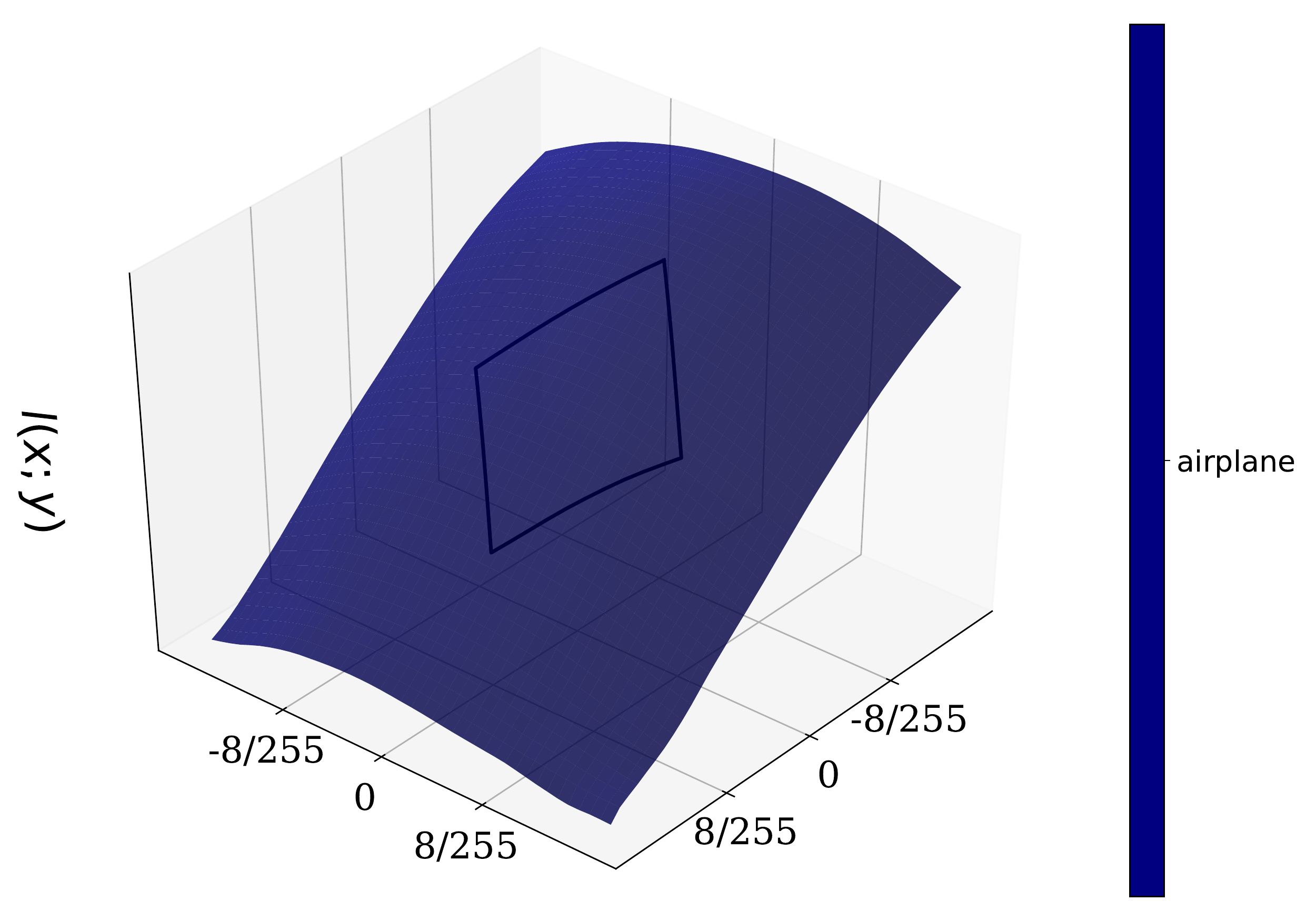}
\caption{airplane}
\end{subfigure}
\begin{subfigure}{0.32\textwidth}
\centering
\includegraphics[width=\linewidth]{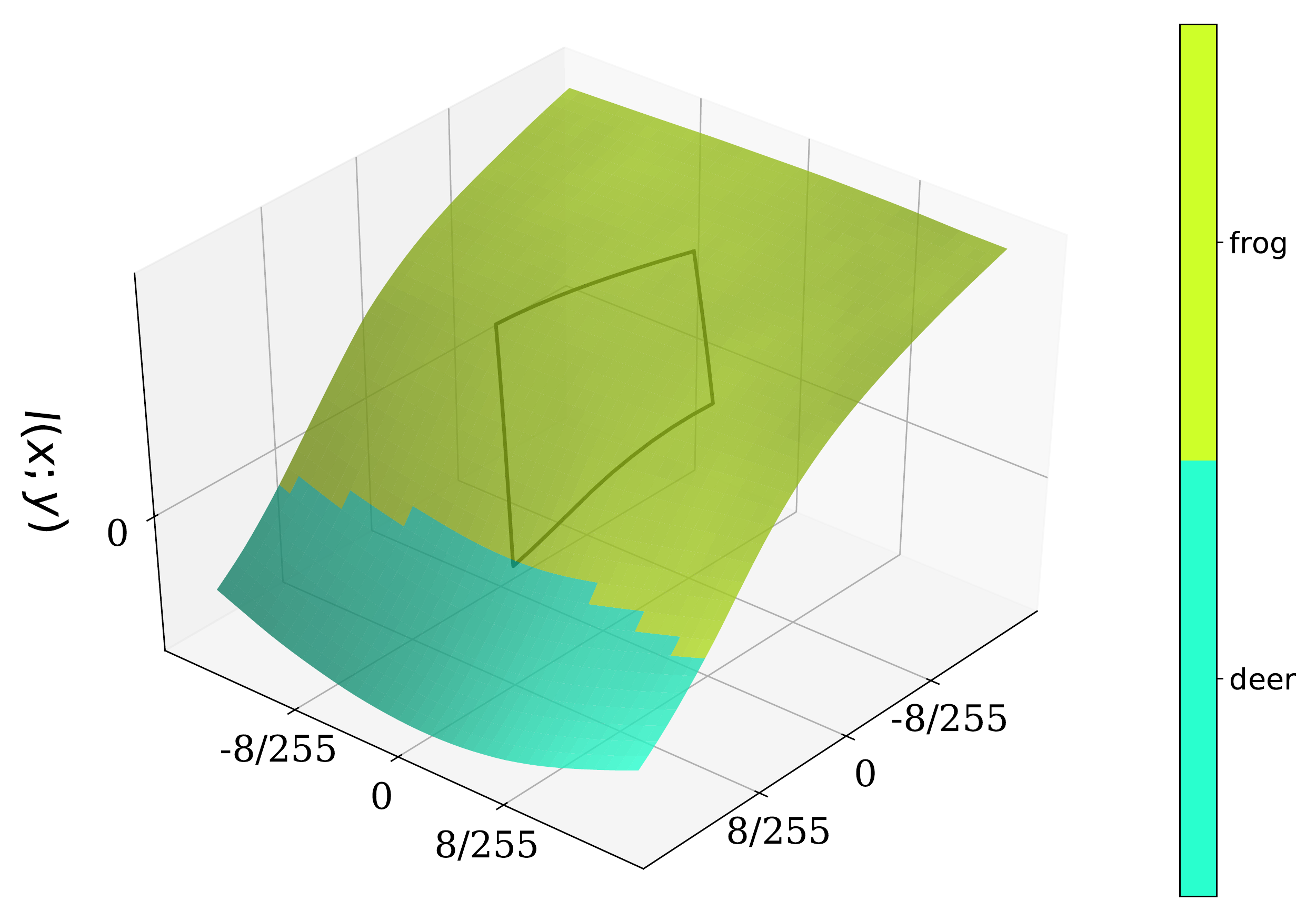}
\caption{frog}
\end{subfigure}
\begin{subfigure}{0.32\textwidth}
\centering
\includegraphics[width=\linewidth]{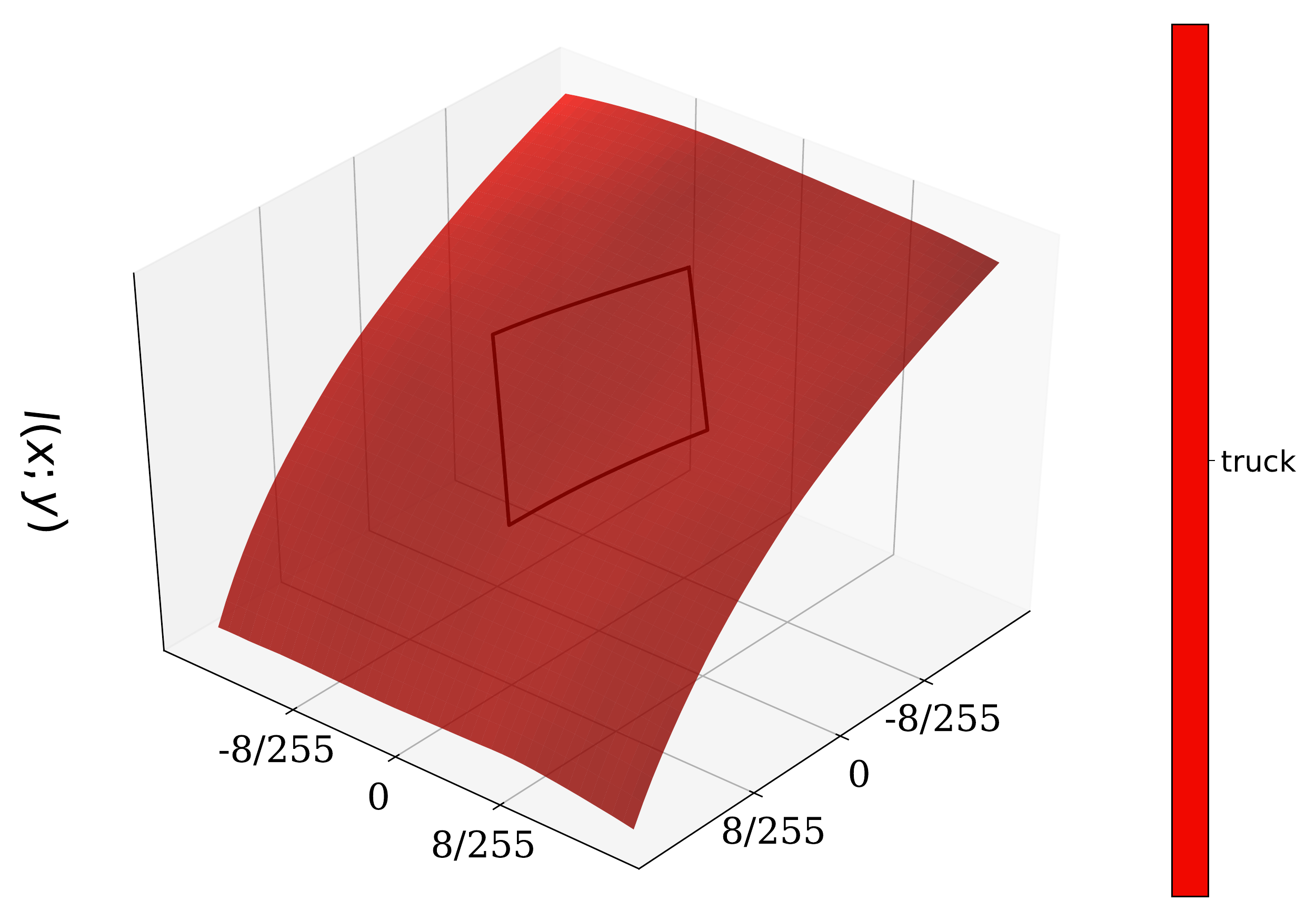}
\caption{truck}
\end{subfigure}
\begin{subfigure}{0.32\textwidth}
\centering
\includegraphics[width=\linewidth]{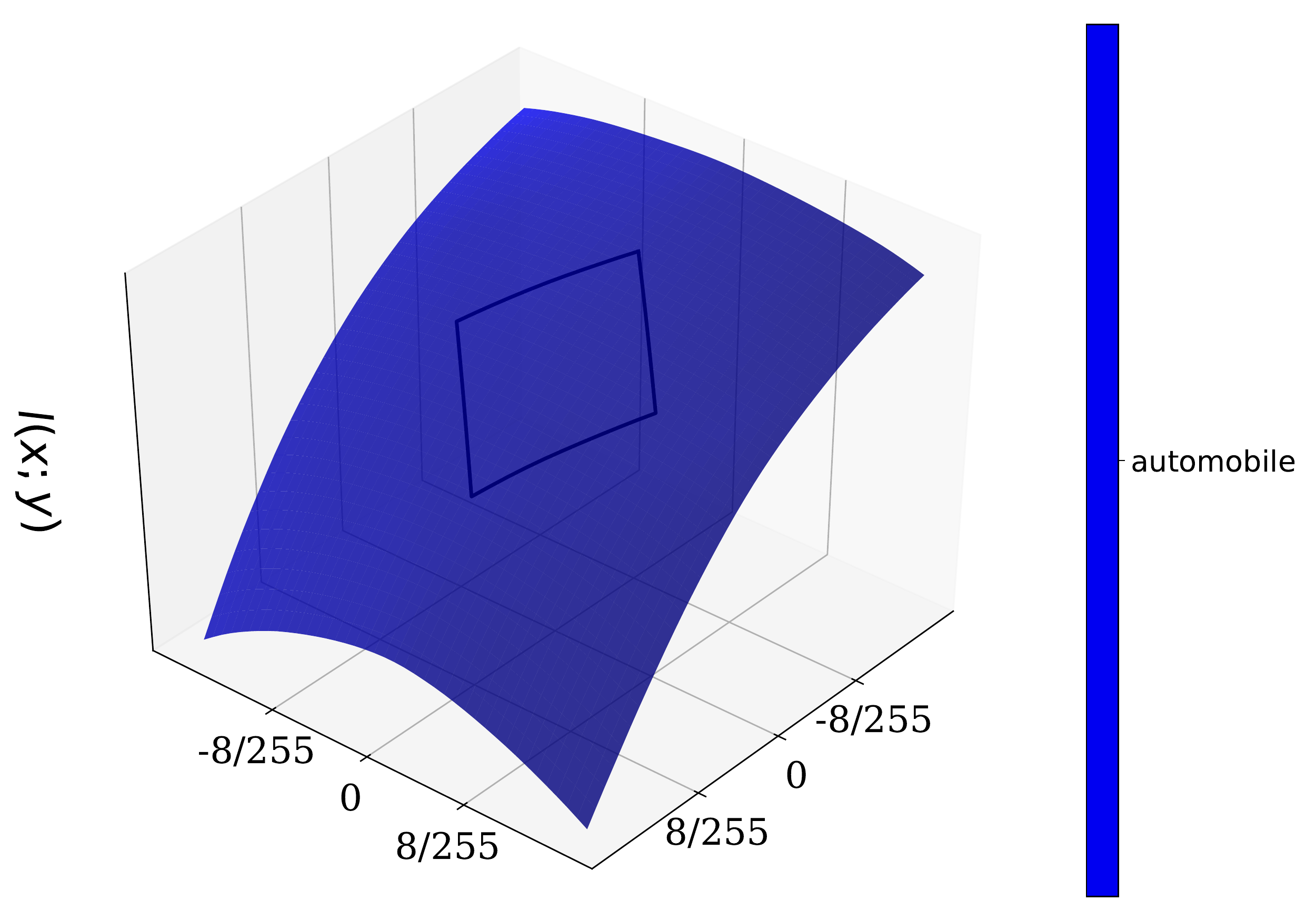}
\caption{automobile}
\end{subfigure}
\begin{subfigure}{0.32\textwidth}
\centering
\includegraphics[width=\linewidth]{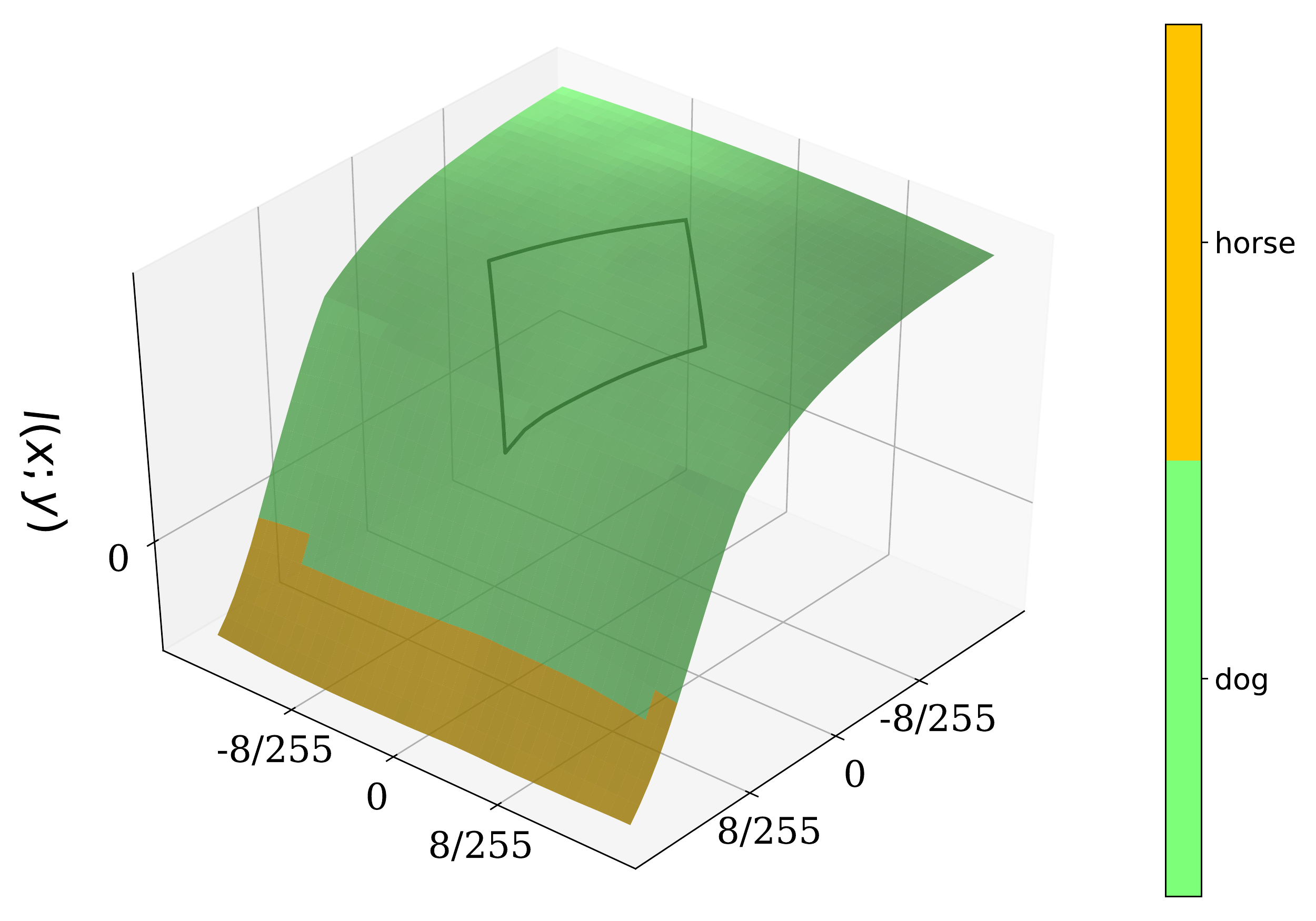}
\caption{dog}
\end{subfigure}
\caption{Loss landscapes around different \cifar test images.
It is generated by varying the input to the model, starting from the original input image toward either the worst attack found using \pgd{40} ($u$ direction) or a random Rademacher direction ($v$ direction). The loss used for these plots is the margin loss $z_y - \max_{i \neq y} z_i$ (i.e., a misclassification occurs when this value falls below zero). The model used is \citeauthor{carmon_unlabeled_2019}'s \wrn-28-10 model trained with additional unlabeled data against \linf norm-bounded perturbations of size $8/255$ on \cifar. The diamond-shape represents the projected \linf ball of size $\epsilon = 8/255$ around the nominal image.}
\label{fig:linf_carmon_landscapes}
\end{figure*}

\autoref{fig:linf_data_landscapes}, \autoref{fig:linf_landscapes} and \autoref{fig:linf_carmon_landscapes} shows the loss landscapes around the first 5 images of the \cifar test set for our largest model trained with additional unlabeled data, our largest model trained without additional unlabeled data and \citeauthor{carmon_unlabeled_2019}'s model trained with additional unlabeled data, respectively.
Most landscapes are smooth and do not exhibit patterns of gradient obfuscation.
Interestingly, the landscapes corresponding to the fifth image (of a dog) are quite similar across all models, with the larger models' landscapes being less smooth (with a cliff).
Overall, it is difficult to interpret these figures further and we rely on the fact that \autoattack and \multitargeted are accurate.
We note that the black-box Square attack~\cite{andriushchenko_square_2019} does not find any misclassified attack that was not already found by \autopgd and \multitargeted.

\end{document}